\documentclass[10pt,twocolumn,letterpaper]{article}

\usepackage[pagenumbers]{cvpr} 

\makeatletter
\robustify\@latex@warning@no@line
\makeatother
\usepackage{authblk}

\usepackage{graphicx}
\usepackage{amsmath}
\usepackage{amssymb}
\usepackage{booktabs}

\usepackage{color}
\usepackage{mathrsfs}
\usepackage{algorithm}
\usepackage{algpseudocode}
\usepackage{multirow}
\usepackage{caption}
\usepackage{subcaption}
\usepackage{diagbox}
\usepackage[accsupp]{axessibility}

\usepackage{xcolor,colortbl}
\definecolor{Gray}{gray}{0.85}
\newcolumntype{a}{>{\columncolor{Gray}}c}

\usepackage[pagebackref,breaklinks,colorlinks]{hyperref}

\usepackage[capitalize]{cleveref}
\crefname{section}{Sec.}{Secs.}
\Crefname{section}{Section}{Sections}
\Crefname{table}{Table}{Tables}
\crefname{table}{Tab.}{Tabs.}

\def\cvprPaperID{6194} 
\def\confName{CVPR}
\def\confYear{2022}

\begin{document}

\title{Not All Tokens Are Equal: Human-centric Visual Analysis via \\Token Clustering  Transformer}

\makeatletter
\renewcommand\AB@affilsepx{ \protect\Affilfont}
\makeatother

\author[1]{Wang Zeng}
\author[2,3]{Sheng Jin}
\author[3]{Wentao Liu}
\author[3]{Chen Qian}
\author[2]{Ping Luo}
\author[4]{Wanli Ouyang}
\author[1]{Xiaogang Wang}

\affil[1]{\small The Chinese University of Hong Kong} 
\affil[2]{The University of Hong Kong}
\affil[3]{SenseTime Research and Tetras.AI}
\affil[4]{The University of Sydney}

\affil[ ]{\authorcr\tt\small \{zengwang@link, xgwang@ee\}.cuhk.edu.hk, \{js20@connect, pluo@cs\}.hku.hk, 
\authorcr\tt\small\{liuwentao, qianchen\}@sensetime.com, wanli.ouyang@sydney.edu.au  }

\maketitle

\begin{abstract}
Vision transformers have achieved great successes in many computer vision tasks. Most methods generate vision tokens by splitting an image into a regular and fixed grid and treating each cell as a token. However, not all regions are equally important in human-centric vision tasks, \eg, the human body needs a fine representation with many tokens, while the image background can be modeled by a few tokens. 
To address this problem, we propose a novel  Vision Transformer, called Token Clustering Transformer (TCFormer), which merges tokens by progressive clustering, where the tokens can be merged from different locations with flexible shapes and sizes. The tokens in TCFormer can not only focus on important areas but also adjust the token shapes to fit the semantic concept and adopt a fine resolution for regions containing critical details, which is beneficial to capturing detailed information.
Extensive experiments show that TCFormer consistently outperforms its counterparts on different challenging human-centric tasks and datasets, including whole-body pose estimation on COCO-WholeBody and 3D human mesh reconstruction on 3DPW. 
Code is available at \url{https://github.com/zengwang430521/TCFormer.git}.

\end{abstract}


\section{Introduction}

\begin{figure}[t]
	\centering
	\includegraphics[width=1\linewidth]{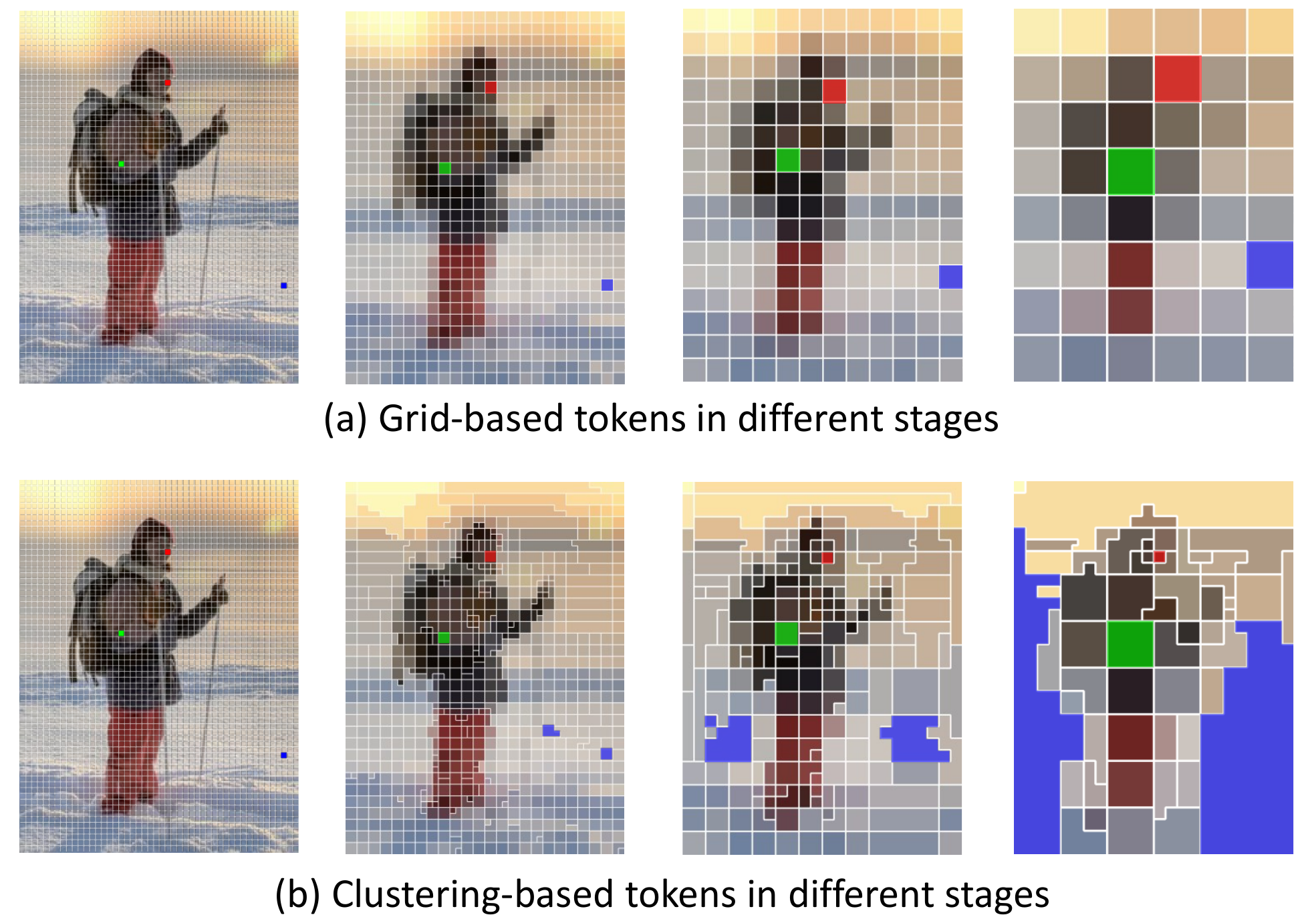}
	\caption{
	Comparisons between vision tokens generated by (a) standard grids and (b) TCFormer. The token regions of different tokens, or the image regions represented by vision tokens, are visualized by different colors. 
	From left to right, different images represent different stages.
	Grid-based tokens treat all regions equally as shown in (a).
	While the tokens in (b) treat image regions dynamically.
	Tokens distribute more densely on the human body. 
	For background regions, a large area is represented by a single token (in blue), while for the regions containing important details, such as the face area, tokens with fine spatial sizes are used (in red). 
	}
	\label{fig:introduction}
\end{figure}

Human-centric tasks~\cite{yu2014spectral,hu2019graph,jin2019multi,wu2021graph,duan2019trb} of computer vision such as face alignment~\cite{zhu2015face,burgos2013robust,wu2018look}, human pose estimation~\cite{toshev2014deeppose,ouyang2014multi,chu2017multi,yang2017learning,jin2020differentiable,liu2018cascaded,li2021human,newell2016stacked,xu2021vipnas,wang2021human}, and 3D human mesh reconstruction~\cite{kanazawa2018end,varol2018bodynet,kolotouros2019learning,zeng20203d} have drawn increasing research attention owing to their broad applications such as action recognition,  virtual reality, and augmented reality.

Inspired by the success of transformers in natural language processing, vision transformers are recently developed to solve human-centric computer vision tasks and achieve state-of-the-art performance~\cite{li2021tokenpose,yuan2021hrformer,mao2021tfpose,yang2020transpose,lin2021mesh}. The properties of transformers such as the long-range attention between image patches are beneficial to model the relationship between different body parts and thus are critical in human-centric visual analysis.

Since the traditional transformers employed a sequence of tokens as input, most existing vision transformers follow this paradigm by dividing an input image into a regular and fixed grid, where each cell (image patch) is treated as a token as shown in Figure~\ref{fig:introduction} (a). The grid-based token generation is simple and achieves great successes in many computer vision tasks~\cite{dosovitskiy2020image,pvt,swin} such as image recognition, object detection, and segmentation.

However, fixed grid based vision tokens are sub-optimal for human-centric visual analysis. 
In human-centric visual analysis, the image regions of the human body are more crucial than the image background, motivating us to represent different image regions by vision tokens with dynamic shape and size
\footnote{
We call the image region represented by a token as the token region and use token location, shape, and size to denote that of its token region.}. 
But the token regions of the grid-based vision tokens are rectangular areas with fixed location, shape and size. 
Uniform vision token distribution is not able to allocate more tokens to important areas.

To solve this problem, we propose a novel vision transformer, named Token Clustering Transformer (TCFormer), which generates tokens by progressive token clustering. 
TCFormer generates tokens dynamically at every stage. 
As shown in Figure~\ref{fig:introduction} (b), it is able to generate tokens with various locations, sizes, and shapes.
Firstly, unlike the grid-based tokens, tokens after clustering are not limited to the regular shape and can focus on important areas \eg, the human body. 
Secondly, TCFormer dynamically generates tokens with appropriate sizes to represent different regions. For the regions full of important details such as the human face, tokens with finer size are allocated. 
In contrast, a single token (\eg, the token in blue in Figure~\ref{fig:introduction} (b)) is used to represent a large area of the background.

In TCFormer, every pixel in the feature map is initialized as a vision token at the first stage, whose token region is the region covered by the pixel.
We progressively merge tokens with similar semantic meanings and obtain different numbers of tokens in different stages.
To this end, we carefully design a Clustering Token Merge (CTM) block. 
Firstly, given tokens from the previous stage, CTM groups them by applying the k-nearest-neighbor based density peaks clustering algorithm~\cite{du2016study} on the token features.
Secondly, the tokens assigned to the same cluster are merged to a single token by averaging the token features.
Finally, the tokens are fed into a transformer block for feature aggregation.
The token region of the merged token is the union of the input token regions.

Aggregation of multi-stage features is proved to be beneficial for human-centric analysis~\cite{sun2019deep,yuan2021hrformer}. Most prior works~\cite{pvt,yuan2021hrformer,swin} transform vision tokens to feature maps and aggregate features in the form of feature maps. However, when transforming our dynamic vision tokens to feature maps, multiple tokens may locate in the same pixel grid, causing the loss of details.
To solve this problem, we propose a Multi-stage Token Aggregation (MTA) head, which is able to preserve image details in all stages in an efficient way. 
Specifically, the MTA head starts from the tokens in the last stage, and then progressively upsamples tokens and aggregates token features from the previous stage, until features in all the stages are aggregated.
The aggregated tokens are in one-to-one correspondence with pixels in the feature maps and are reshaped to the feature maps for subsequent processing.

We summarize our contributions as follows.
 \begin{enumerate}
	\item[$\bullet$] 
	We propose a Token Clustering Transformer (TCFormer), which generates vision tokens of various locations, sizes, and shapes for each image by progressive clustering and merging tokens. To the best of our knowledge, it is the first time that clustering is used for dynamic token generation.

	\item[$\bullet$] We propose a Multi-stage Token Aggregation (MTA) head to aggregate token features in multiple stages, reserving detailed information in all stages efficiently.
	
	\item[$\bullet$] Extensive experiments show that TCFormer consistently outperforms its counterparts on different challenging human-centric tasks and datasets, including whole-body pose estimation on COCO-WholeBody and 3D human mesh reconstruction on 3DPW.
\end{enumerate}

\vspace{+5pt}

\begin{figure*}[t]
	\centering
	\includegraphics[width=1\linewidth]{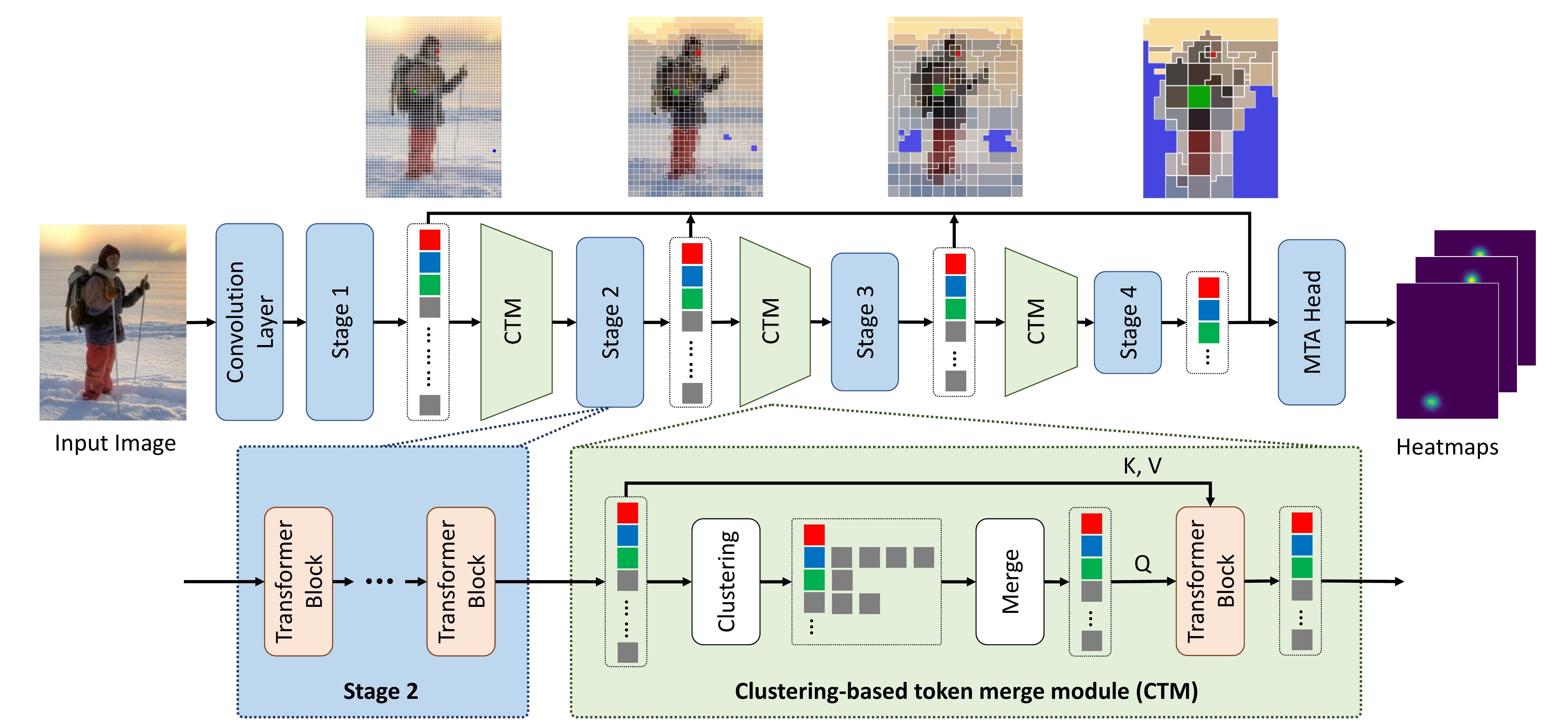}
	\vspace{-12pt}
	\caption{
	Overview of Token Clustering Transformer (TCFormer). 
	Given an input image, feature map is extracted with a single convolution layer and pixels in the feature map is regarded as initial vision tokens. These initial vision tokens are processed in a multi-stage manner and every stage is composed of multiple stacked transformer blocks.
    At the end of every stage, vision tokens are merged with a Clustering-based Token Merge (CTM) block to generate tokens for the next stage. 
    Finally, all vision tokens are fed into the Multi-stage Token Aggregation (MTA) head for predicting heatmaps.
	In the CTM block, input tokens are first clustered and then merged by weighted averaging.
	Merged tokens and original tokens are then fed into a transformer block for better feature aggregation. 
	}
	\label{fig:framework}
	\vspace{-5pt}
\end{figure*}

\section{Related Works}

\subsection{Transformers in Human-Centric Vision Tasks}
Modeling the interaction between human and environment, and the relationships between body parts is a key point in human-centric vision tasks. With global receptive fields, transformers have achieved great success recently in 2d pose estimation~\cite{li2021tokenpose,yuan2021hrformer,mao2021tfpose,yang2020transpose}, 3d pose estimation~\cite{zheng20213d,hampali2021handsformer,huang2020hot} and 3d mesh reconstruction~\cite{lin2021mesh,lin2021end}. 
Prior works can be divided into two categories, \ie refining image features and aggregating part-wise features.

The first kind of method applies transformers to extract better image features. For example, TransPose~\cite{yang2020transpose} refines the image features extracted by CNN with extra transformer encoder layers. HRFormer~\cite{yuan2021hrformer} applies transformer blocks in HRNet~\cite{sun2019deep} structure.
Our proposed TCFormer also belongs to the first kind.

The second kind of method applies transformers to aggregate features for different parts. 
For example, TFPose~\cite{mao2021tfpose} and TokenPose~\cite{li2021tokenpose} use stacked transformers to generate keypoint-wise features, from which the keypoint location is predicted by regressing coordinates and heatmaps respectively.
Mesh Graphormer~\cite{lin2021mesh} designs a head with transformers and graph convolution layers. The head aggregates joint-wise and vertex-wise features from image feature maps and predicts the 3D location of human joints and mesh vertices.

Most prior works in both categories generate vision tokens with fixed grid token region, which is sub-optimal for human-centric tasks. 
In contrast, the token regions of TCFormer are learned automatically rather than using a hand-crafted design. The learned token regions are flexible in location, shape and size according to the semantic meanings.

\subsection{Dynamic Token Generation}
Recently, there are increasing exploration about dynamic token generation in vision transformers~\cite{rao2021dynamicvit,wang2021not,yue2021vision,wang2021pnp}.
DynamicViT~\cite{rao2021dynamicvit} and PnP-DETR~\cite{wang2021pnp} pick up important tokens by predicting token-level scores.
For unimportant tokens, DynamicViT simply discards them, while PnP-DETR represents them with sparse feature vectors.
DVT~\cite{wang2021not} builds vision tokens with grids in different resolutions according to the difficulty of classification. 
PS-ViT~\cite{yue2021vision} builds tokens with a fixed grid size and progressively adjusts the grid centers during processing.

The methods mentioned above are all variants of grid-based token generation. They modify the number, resolution, and centers of grids specifically. 
In contrast, the token regions of our proposed TCFormer are not restricted by grid structure and are more flexible in three aspects, \ie location, shape, and size.
Firstly, we assign image regions to certain vision tokens based on the semantic similarity instead of the spatial proximity. 
Secondly, the image region of a token is not restricted to a rectangular shape. Regions with the same semantic meaning, even if they are non-adjacent, can be represented by a single token.
Thirdly, vision tokens in the same stage may have different sizes. 
For example, in Figure~\ref{fig:introduction} (b), the token for the background presents a large region, while the token for the human face only presents a small region. Such property is helpful for retaining important details among all the stages of TCFormer.

\subsection{Clustering for Feature Aggregation}
Clustering-based feature aggregation methods are wildly explored in research areas of point clouds~\cite{qi2017pointnet++} and graph representations~\cite{ying2018hierarchical}.
PointNet++~\cite{qi2017pointnet++} downsamples point clouds by farthest point sampling and then aggregates features of the k-nearest neighbors.
DIFFPOOL~\cite{ying2018hierarchical} predicts soft cluster assignment matrices for an input graph, which are used to hierarchically coarsen the graph. 

These methods are specially designed for point cloud and graph data, and cannot be directly applied to the image-based vision transformers.
In contrast, we cluster the tokens in the different stages for image-based human-centric vision tasks.
To the best of our knowledge, it is the first time that clustering is utilized for vision token generation.

\section{Method}

\subsection{Overview Architecture}
As shown in Figure \ref{fig:framework}, our proposed Token Clustering Transformer (TCFormer) follows the popular multi-stage architecture.
TCFormer consists of 4 hierarchical stages and a Multi-stage Token Aggregation (MTA) head. 
Each stage contains several stacked transformer blocks.
Between two adjacent stages, a Clustering-based Token Merge (CTM) block is inserted to merge tokens and generate tokens for the next stage.
MTA head aggregates token features from all stages and outputs the final heatmaps.

\subsection{Transformer Block}
\label{sec:transformerblock}

\begin{figure}[t]
	\centering
	\includegraphics[width=1\linewidth]{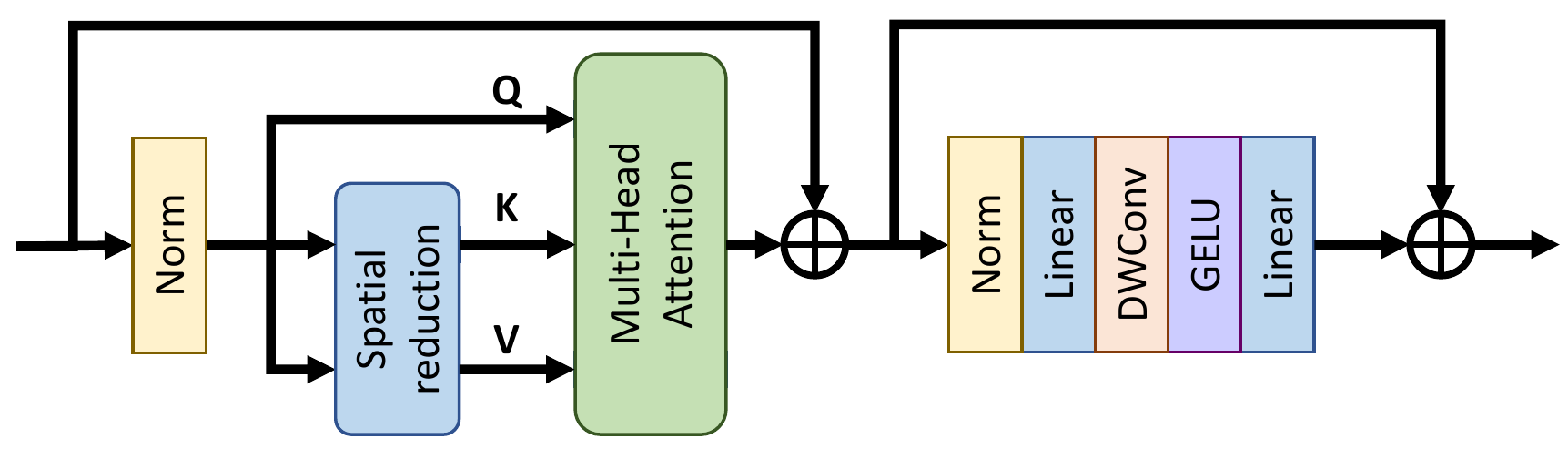}
	\caption{
	Structure of the transformer block in TCFormer.
	A spatial reduction module is added before the multi-head attention module to reduce the computational complexity. 
	A depth-wise convolutional layer is added after the attention module to capture the local information.
	}
	\label{fig:block}
	\vspace{-2mm}
\end{figure}

Figure \ref{fig:block} shows the transformer block used in TCFormer.
Following~\cite{pvt},  a spatial reduction layer is used to reduce the computational complexity.
The spatial reduction layer first transforms the vision tokens to feature maps and then reduces the feature map resolution with a strided convolutional layer. 
The pixels in the processed feature maps, which are much fewer than the vision tokens, are fed into the multi-head attention module as keys and values.
The multi-head attention module aggregates feature between tokens.
Inspired by \cite{localvit,cvt,yuan2021hrformer}, we utilize a depth-wise convolutional layer to capture local feature and positional information, and remove explicit positional embedding.

\subsection{Clustering-based Token Merge (CTM) Block}
\label{sec:CTM}
As shown in Figure \ref{fig:framework}, our Clustering-based Token Merge (CTM) block has two processes, \ie \emph{token clustering} and \emph{feature merging}.
We apply \emph{token clustering} to group vision tokens into a certain number of clusters using token features, and then apply \emph{feature merging} to merge the tokens in the same cluster into a single token.

\textbf{Token Clustering.}
In the token clustering process, we utilize a variant of k-nearest neighbor based density peaks clustering algorithm (DPC-KNN)~\cite{du2016study}. 

Given a set of tokens $X$, we compute the local density $\rho$ of each token according to its k-nearest neighbors:
\begin{equation}
\rho_{i}=\exp \left(-\frac{1}{k} \sum_{x_{j} \in \mathrm{KNN}\left({x}_{i}\right)} 
\left\|x_i-x_j\right\|_{2}^{2},
\right),
\label{eq:rho}
\end{equation}
where $\mathrm{KNN}\left({x}_{i}\right)$ denotes the k-nearest neighbors of a token $i$. $x_i$ and $x_j$ are their corresponding token features.

Then, for each token, we compute the distance indicator as the minimal distance between it and any other token with higher local density. 
For the token with the highest local density, its indicator is set as the maximal distance between it and any other tokens. 
\begin{equation}
\delta_{i}=\left\{\begin{array}{l}
\min _{j: \rho_{j}>\rho_{i}} \left\|x_i-x_j\right\|_{2}, \text { if } \exists j \text { s.t. } \rho_{j}>\rho_{i} \\
\max _{j} \left\|x_i-x_j\right\|_{2}, \text { otherwise }
\end{array}\right.
\label{eq:delta}
\end{equation}
where $\delta_{i}$ is the distance indicator and $\rho_{i}$ is the local density.

We combine the local density and the distance indicator to get the score of each token as $\rho_i \times \delta_i$. Higher scores mean higher potential to be cluster centers. 
We determine cluster centers by selecting the tokens with the highest scores, and then assign other tokens to the nearest cluster center according to the feature distances.

\textbf{Feature Merging.}
For token features merging, an intuitive method is to average the token features in the cluster directly. However, even for the tokens with similar semantic meanings, the importance is not totally the same.

Inspired by \cite{rao2021dynamicvit}, we introduce an importance score $P$ to explicitly represent the importance of each token, which is estimated from the token features.
The token features are averaged with the guidance of the token importance:
\begin{equation}
y_{i}=\frac {\sum_{j \in C_i} e^{p_j} x_j}
            {\sum_{j \in C_i} e^{p_j}} ,
\label{eq:average}
\end{equation}
where $C_i$ means the set of the $i$-th cluster, $x_j$ and $p_j$ are the original token features and the corresponding importance score respectively, and $y_i$ is the features of the merged token. The token region of the merged token is the union of 
the original token regions.

As shown in Figure~\ref{fig:framework}, the merged tokens are fed into a transformer block as queries $Q$, and the original tokens are used as keys $K$ and values $V$. 
To ensure that important tokens contribute more to the output, the importance score $P$ is added to the attention weight as follows:
\begin{equation}
\operatorname{Attention}(Q, K, V)=\operatorname{softmax}\left(\frac{Q K^{T}}{\sqrt{d_{k}}}+P\right) V,
\label{eq:attn}
\end{equation}
where $d_k$ is the channel number of the queries. We omit the multi-head setting and the spatial reduction layer here for clarity.
Introducing the token importance score equips our CTM block with the capability to focus on the critical image features when merging vision tokens.

We adopt an efficient GPU implementation of CTM block. The clustering and feature merging parts only cost $9.4\%$ of the forward time of TCFormer.

\begin{figure}[t]
	\centering
	\includegraphics[width=1\linewidth]{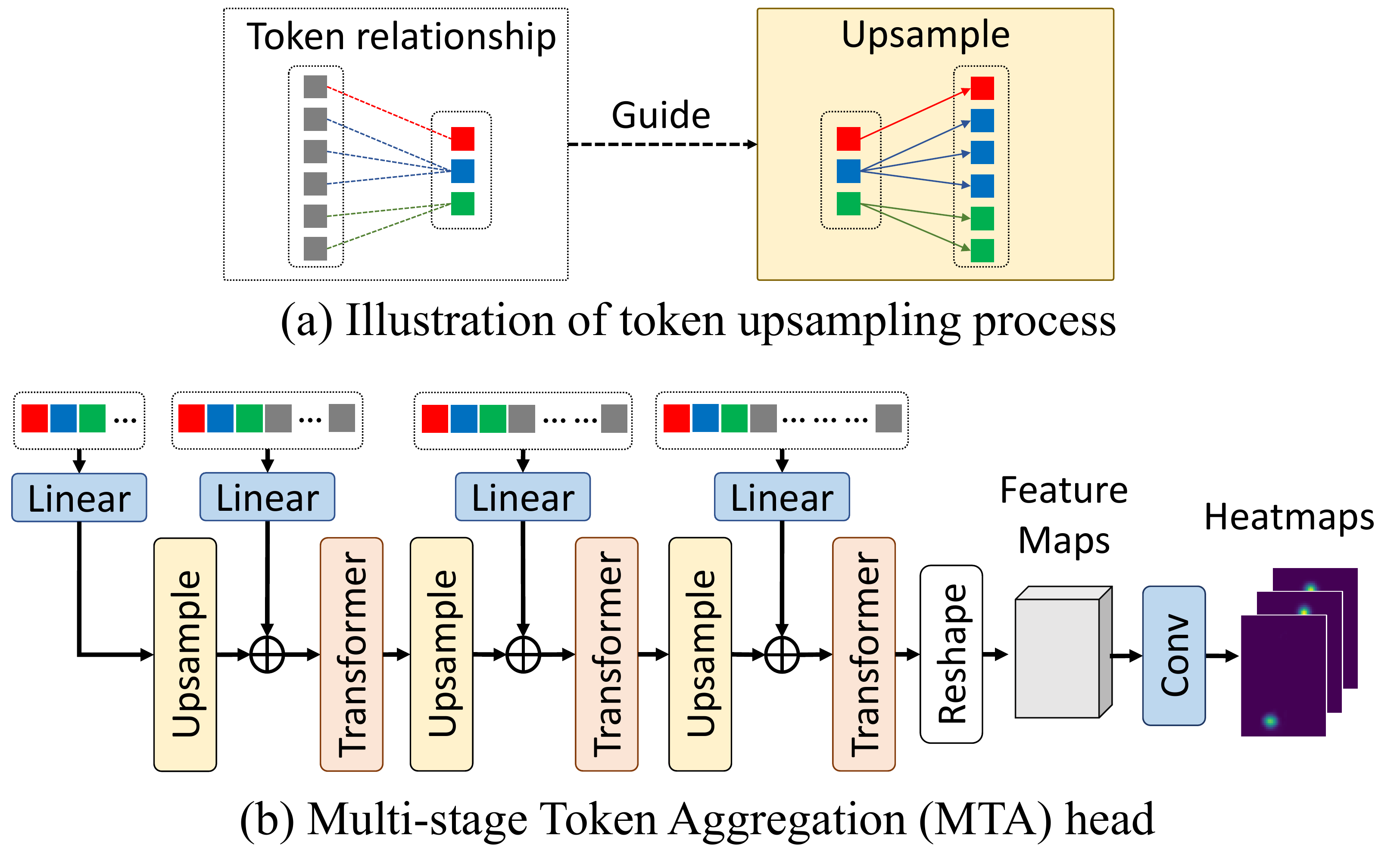}
	\vspace{-15pt}
	\caption{
	Multi-stage Token Aggregation (MTA) head. 
	(a) 
	In token upsampling process, we use the recorded token relationship to copy the merged token features to the corresponding upsampled tokens.
	(b) MTA head starts from the last stage and progressively aggregates features by stacked upsampling processes and transformer blocks. 
	The final tokens are in one-to-one correspondence with the pixels in feature maps and reshaped to the feature maps for further processing.
	}
	\vspace{-10pt}
	\label{fig:head}
\end{figure}

\subsection{Multi-stage Token Aggregation (MTA) Head}
\label{sec:MTA}
Prior works~\cite{sun2019deep,yuan2021hrformer} prove the benefits of feature aggregation in multiple stages for human-centric vision tasks.
In order to aggregate features, we propose a transformer-based Multi-stage Token Aggregation (MTA) head, which is able to maintain details in all the stages.

Figure \ref{fig:head} (a) shows the token upsampling process.
During the token merging process (Section~\ref{sec:CTM}), each token is assigned to a cluster and each cluster is represented by a single merged token. We record the relationship between the original tokens and merged tokens. In the token upsampling process, we use the recorded information to copy the merged token features to the corresponding upsampled tokens.
As shown in Figure \ref{fig:head} (b), after the token upsampling, MTA head adds the token features in the previous stage to the upsampled vision tokens. The vision tokens are then processed by a transformer block. Such processing is executed progressively until all vision tokens are aggregated. 
The final tokens, whose token region is a single pixel in the high-resolution feature map, can be easily reshaped to feature maps for further processing.

\begin{figure}[t]
	\centering
	\includegraphics[width=1\linewidth]{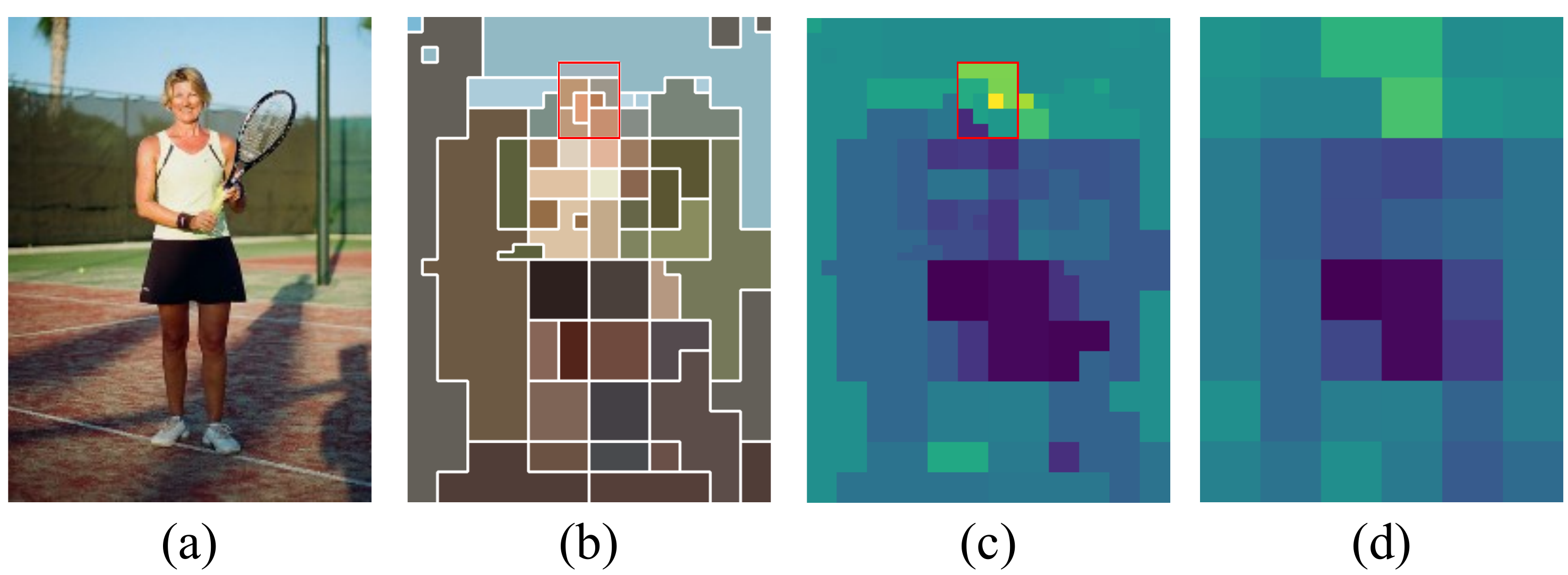}
	\vspace{-5mm}
	\caption{
	A typical example of vision tokens generated by TCFormer.
	(a) Input image. 
	(b) Generated vision tokens in stage 4. Regions with the same color are represented by the same vision token. For the regions with details, such as the face area,  vision tokens are with fine sizes.
    (c)(d)High resolution and low-resolution feature maps transformed from vision tokens.
	Transforming vision tokens to low-resolution feature maps causes detail loss. 
	}
	\vspace{-3mm}
	\label{fig:analysis}
\end{figure}

\textbf{Why not CNN?}
Most prior works~\cite{pvt,yuan2021hrformer,swin} transform vision tokens to feature maps first and aggregate multi-stage features with convolutional layers.
However, as shown in Figure~\ref{fig:analysis}, vision tokens for the regions with important details, such as the human face, are in small size. Transforming vision tokens to low-resolution feature maps involves feature averaging of these tokens, which leads to the loss of details. 
Such loss can be avoided by transforming vision tokens in all stages to high-resolution feature maps, but it will lead to unacceptable complexity and memory cost.
Our MTA head, which aggregates features in the form of tokens and transforms the final tokens to high-resolution feature maps at the end, preserves image details in all stages with relatively low complexity.

\subsection{Implementation Details}

\textbf{Transform between Vision Tokens and Feature Maps.}
The convolutional process is non-trivial for irregular tokens. We need to transform tokens to feature maps before this process and perform the inverse transform after that.
We regard every pixel in feature maps as a representation of a rectangular grid region. 
When transforming vision tokens to feature maps, we first find the related vision tokens for every pixel, \ie the vision tokens whose token region overlaps with the grid region of the pixel. Then we average the token features according to the overlapped areas to get the image feature of every pixel.
When transforming feature maps to vision tokens, we directly set the token features as the average image features in the token region.

\textbf{Cluster Numbers.}
In the CTM block, the cluster number can be any value smaller than the input token number. 
In this work, we set the cluster number to a quarter of the input token number to keep the token numbers in every stage the same as those of the prior works~\cite{pvt,swin}.

\section{Experiments}

\begin{table*}[t]
	\caption{ OKS-based Average Precision (AP) and Average Recall (AR) on the COCO-WholeBody V1.0 dataset. The baseline results are from MMPose~\cite{mmpose2020}. `*' indicates multi-scale testing.
	ZoomNet$^{\dagger}$ is trained with the COCO-WholeBody V0.5 training set.
	}
	\vspace{-15pt}
	\label{tab:wholebody}
	\begin{center}
    \scalebox{0.94}{
		\begin{tabular}{c|c|cc|cc|cc|cc|aa}
			\hline
			 \multirow{2}{*}{Method} & Resolution &  \multicolumn{2}{c|}{body}  & \multicolumn{2}{c|}{foot}  & \multicolumn{2}{c|}{face}  & \multicolumn{2}{c|}{hand} & \multicolumn{2}{c}{\cellcolor{Gray}whole-body} \\
			\cline{3-12}
			& ~ &  AP     & AR     & AP   & AR     &  AP  & AR     & AP    & AR   &  AP     & AR  \\
			\hline
            SN$^{*}$~\cite{hidalgo2019single} & $480\times480$ & 0.427 & 0.583 & 0.099 & 0.369 & 0.649 & 0.697 & 0.408 & 0.580 & 0.327 & 0.456 \\ 
            OpenPose~\cite{cao2018openpose} & $480\times480$ & 0.563 & 0.612 & 0.532 & 0.645 & 0.765 & 0.840 & 0.386 & 0.433 & 0.442 & 0.523 \\ 
			PAF$^{*}$~\cite{cao2017realtime} & $480\times480$  & 0.381 & 0.526 & 0.053 & 0.278 & 0.655 & 0.701 & 0.359 & 0.528 & 0.295 & 0.405 \\ 
            AE~\cite{newell2017associative}+HRNet-w48~\cite{sun2019deep} & $512\times512$ & 0.592	& 0.686& 	0.443& 	0.595& 	0.619& 	0.674& 	0.347& 	0.438& 	0.422& 	0.532 \\ 
             HigherHRNet-w48~\cite{cheng2020higherhrnet} & $512\times512$ & 0.630 &	0.706 &	0.440 &	0.573 &	0.730 &	0.777 &	0.389 &	0.477 &	0.487 &	0.574 \\ 
            \hline
             ZoomNet$^{\dagger}$~\cite{jin2020whole} & $384\times288$ & 0.743 & 0.802 &  0.798 & 0.869 & 0.623 & 0.701 & 0.401 & 0.498 & 0.541 & 0.658 \\ 
            SBL-Res50~\cite{xiao2018simple}  & $256\times192$ & 0.652 &	0.739 &	0.614 &	0.746 &	0.608 &	0.716 &	0.460 &	0.584 &	0.520 &	0.633 \\
            SBL-Res101~\cite{xiao2018simple}  & $256\times192$ & 0.670 &	0.754 &	0.640 &	0.767 &	0.611 &	0.723 &	0.463 &	0.589 &	0.533 &	0.647 \\
            SBL-Res152~\cite{xiao2018simple}   & $256\times192$ & 0.682 &	0.764 &	0.662 &	0.788 &	0.624 &	0.728 &	0.482 &	0.606 &	0.548 &	0.661 \\
			HRNet-w32~\cite{sun2019deep} & $256\times192$  & 0.700 &	0.746 &	0.567 &	0.645 &	0.637 &	0.688 &	0.473 &	0.546 &	0.553 &	0.626 \\ 
            \hline
			 TCFormer w/o CTM  & $256\times192$ & 0.667 & 0.749&	0.562& 0.695 &	0.617 & 0.621 &	0.479& 0.590 &	0.535 & 0.639 \\ 
			 TCFormer w/o MTA Head & $256\times192$ &0.679 & 0.761&0.658 &0.780 &0.634 &0.732 &0.499 &0.619 &0.553 &0.662\\ 
			 TCFormer (Ours) & $256\times192$ & 0.691 & 0.770 & 0.698 & 0.813 & 0.649 & 0.746 & 0.535 & 0.650 & \textbf{0.572} & \textbf{0.678} \\ 
			\hline
		\end{tabular}
		}
	\end{center}
	\vspace{-10pt}
\end{table*}

\subsection{2D Whole-body Pose Estimation}
\label{sec:human_pose}
Whole-body pose estimation targets at localizing fine-grained keypoints on the entire human body including the face, the hands, and the feet, which requires the ability to capture detailed information.

\textbf{Settings.}
We conduct experiments on COCO-WholeBody V1.0 dataset~\cite{jin2020whole}. COCO-WholeBody dataset is built upon the popular COCO dataset~\cite{lin2014microsoft} with additional whole-body pose annotations. The full pose contains 133 keypoints, including 17 for the body, 6 for the feet, 68 for the face, and 42 for the hands. Following~\cite{lin2014microsoft,jin2020whole}, we use OKS-based Average Precision (AP) and Average Recall (AR) for evaluation.

We follow most of the default training and evaluation settings of mmpose~\cite{mmpose2020} and replace Adam~\cite{kingma2014adam} with AdamW~\cite{loshchilov2017decoupled} with momentum of $0.9$ and weight decay of $1 \times 10^{-2}$.

\textbf{Results.} 
Table~\ref{tab:wholebody} shows the experimental results comparing TCFormer with the state-of-the-art models on COCO-WholeBody V1.0 dataset~\cite{jin2020whole}. 
The results show that the whole-body pose estimation accuracy of TCFormer (57.2\% AP and 67.8\% AR) is higher than those of the state-of-the-art top-down methods, \eg HRNet~\cite{sun2019deep}, by a large margin.

The size of the hand in the input image is relatively small, which makes the estimation of hand keypoint extremely difficult and heavily reliant on the capability of the model in capturing details. 
As shown in Table~\ref{tab:wholebody}, most models achieve much lower performance on the hand than the other parts.
Our TCFormer achieves a large gain in hand keypoint estimation, \ie $6.2\%$ AP higher than HRNet-w32~\cite{sun2019deep} and $5.3\%$ AP higher than SBL-Res152~\cite{xiao2018simple}, which demonstrates the excellent capability of TCFormer in capturing critical image details with small sizes.

\subsection{3D Human Mesh Reconstruction}

Prior works of 3D human mesh reconstruction can be divided into model-based~\cite{kanazawa2018end,kolotouros2019convolutional,kolotouros2019learning,kolotouros2019learning,zanfir2020weakly,joo2020exemplar} and model-free methods~\cite{lin2021end,lin2021mesh,choi2020pose2mesh,moon2020i2l,zeng20203d}. 
We build a model-based method by combining TCFormer and HMR head~\cite{kanazawa2018end}.

\begin{table}[t]
\footnotesize
\caption{ Mean Per Joint Position Error (MPJPE) and the error after Procrustes  alignment (PA-MPJPE) on 3DPW and Human3.6M dataset. $\downarrow$ denotes lower is better.
Our proposed method achieves results competitive with the state-of-the-art models. 
``*'' means training without using 3DPW.
}
\vspace{-15pt}
\label{tab:mesh}
\begin{center}
    \scalebox{0.94}{
	\begin{tabular}{c|cc|cc}
		\hline
		\multirow{2}{*}{Method} &  \multicolumn{2}{c|}{3DPW}  & \multicolumn{2}{c}{Human3.6M} \\
		~ & PA-MPJPE$\downarrow$     & MPJPE$\downarrow$     & PA-MPJPE$\downarrow$     & MPJPE$\downarrow$ \\
		\hline
		HMR~\cite{kanazawa2018end}* & 76.7 & 130.0 & 56.8 & 88 \\
		SPIN~\cite{kolotouros2019learning}* & 59.2 & 96.9 & \textbf{41.1} & 62.5 \\
		Zanfir \etal~\cite{zanfir2020weakly} & 57.1 & 90.0  & - & - \\
		EFT~\cite{joo2020exemplar} & 52.2 & -  & 43.8 & - \\
	    DSR~\cite{dwivedi2021learning}& 51.7 & 85.7  & 41.4 & \textbf{62.0} \\
		\hline
	    TCFormer (Ours) & \textbf{49.3} & \textbf{80.6} & 42.8& 62.9 \\ \hline
	\end{tabular}
	}
\end{center}
\vspace{-10pt}
\end{table}

\begin{table*}[tb]
\centering
\caption{
NME on WFLW \texttt{Test} and $6$ subsets: large pose (LP), expression (Expr.), illumination (Illu.), make-up (Mu.), occlusion (Occu.), and Blur.  $\downarrow$ means lower is better.
LAB~\cite{wu2018look} relies on extra boundary information (B). PDB~\cite{feng2018wing} uses stronger data augmentation (DA). 
}
\vspace{-2mm}
\label{table:comparison_wflw_testset}

\small
\begin{tabular}{l|l|r|r|r|r|r|r|r }
\hline \noalign{\smallskip}
Method   & Backbone & Test $\downarrow$ & LP  $\downarrow$& Expr. $\downarrow$ & Illu. $\downarrow$ & Mu. $\downarrow$ & Occu.  $\downarrow$& Blur $\downarrow$\\
\hline

\hline
ESR \cite{cao2014face}& - & $11.13$ & $25.88$ & $11.47$ & $10.49$ & $11.05$ & $13.75$ & $12.20$\\
SDM \cite{xiong2013supervised}& - &$10.29$ & $24.10$ & $11.45$ & $9.32$ & $9.38$ & $13.03$ & $11.28$\\
CFSS \cite{zhu2015face}& - &$9.07$ & $21.36$ & $10.09$ & $8.30$ & $8.74$ & $11.76$ & $9.96$\\
DVLN \cite{wu2017leveraging} & VGG-16&$6.08$ & $11.54$ & $6.78$ & $5.73$ & $5.98$ & $7.33$ & $6.88$\\
\hline
HRNetV2 \cite{wang2020deep} & HRNetV$2$-W$18$ & $4.60$ & $7.94$ & $4.85$ & $4.55$ & $4.29$ & $5.44$ & $5.42$\\
TCFormer (Ours) & TCFormer-Light & \textbf{4.28} & \textbf{7.27} & \textbf{4.56} & \textbf{4.18} & \textbf{4.27} & \textbf{5.18}  &  \textbf{4.87}\\
\hline
\hline
\multicolumn{3}{l}{
\emph {Model trained with \texttt{extra} info.}}\\
\hline
LAB (w/ B)~\cite{wu2018look}& Hourglass & $5.27$ & $10.24$ & $5.51$ & $5.23$ & $5.15$ & $6.79$ & $6.32$\\
PDB (w/ DA)~\cite{feng2018wing}& ResNet-$50$ & $5.11$ & $8.75$ & $5.36$ & $4.93$ & $5.41$ & $6.37$ & $5.81$\\
\hline
\end{tabular} 
\end{table*}

\textbf{Dataset.}
We evaluate our model on 3DPW dataset~\cite{von2018recovering} and Human3.6M dataset~\cite{ionescu2013human3}. 3DPW is composed of more than 51K frames with accurate 3D mesh in the wild, and Human3.6M is a large-scale indoor dataset for 3D human pose estimation.
Mean Per Joint Position Error (MPJPE) and the error after Procrustes alignment (PA-MPJPE) are reported.
Following the setting of \cite{kanazawa2018end}, we train our model with a mixture of datasets, including Human3.6M~\cite{ionescu2013human3}, MPI-INF-3DHP~\cite{mehta2017monocular}, LSP~\cite{Johnson10}, LSP-Extended~\cite{johnson2011learning}, MPII~\cite{andriluka20142d} and COCO~\cite{lin2014microsoft}. For fair comparisons with the recent methods~\cite{joo2020exemplar,dwivedi2021learning}, we also use the training set of 3DPW and the pseudo-label provided by ~\cite{joo2020exemplar}.

\textbf{Settings.} 
We crop the image with the ground truth bounding box and resize it to the resolution of $224\times224$. During training, we augment the data with random scaling, random rotation, random flipping, and random color jittering. 
The model is trained with 8 GPUs with a batch size of 32 in each GPU for 80 epochs. We use Adam optimizer~\cite{kingma2014adam} with the learning rate of $2.5 \times 10^{-4}$ and do not use learning rate decay. 
The settings are the same as that in~\cite{joo2020exemplar,dwivedi2021learning}.

\textbf{Results.} 
We compare TCFormer with the state-of-the-art SMPL-based methods and show the results in Table~\ref{tab:mesh}. TCFormer outperforms most of the prior works~\cite{kanazawa2018end,kolotouros2019learning,zanfir2020weakly,joo2020exemplar}
with similar structures and model complexity.
TCFormer even obtains competitive results compared with the DSR method that uses extra dense supervision~\cite{dwivedi2021learning}. 
The results of human mesh estimation further validate the effectiveness of TCFormer in capturing important image features. It works well not only on the dense prediction task, but also on the global feature based regression task.

\subsection{2D Face Keypoint Localization}

\textbf{Settings.}
We perform experiments on WFLW~\cite{wu2018look} dataset, which consists of $7,500$ training and $2,500$ testing images with $98$ landmarks. The evaluation is conducted on the test set and several subsets: large pose, expression, illumination, make-up, occlusion, and blur.
We use the normalized mean error (NME) as the evaluation metric and the inter-ocular distance for normalization. 
We apply the same training and evaluation settings as that of~\cite{wang2020deep}.
For fair comparisons, we use a lightweight version of TCFormer (TCFormer-Light) with similar model complexity as~\cite{wang2020deep}.

\begin{figure*}[t]
	\centering
	\includegraphics[width=0.75\linewidth]{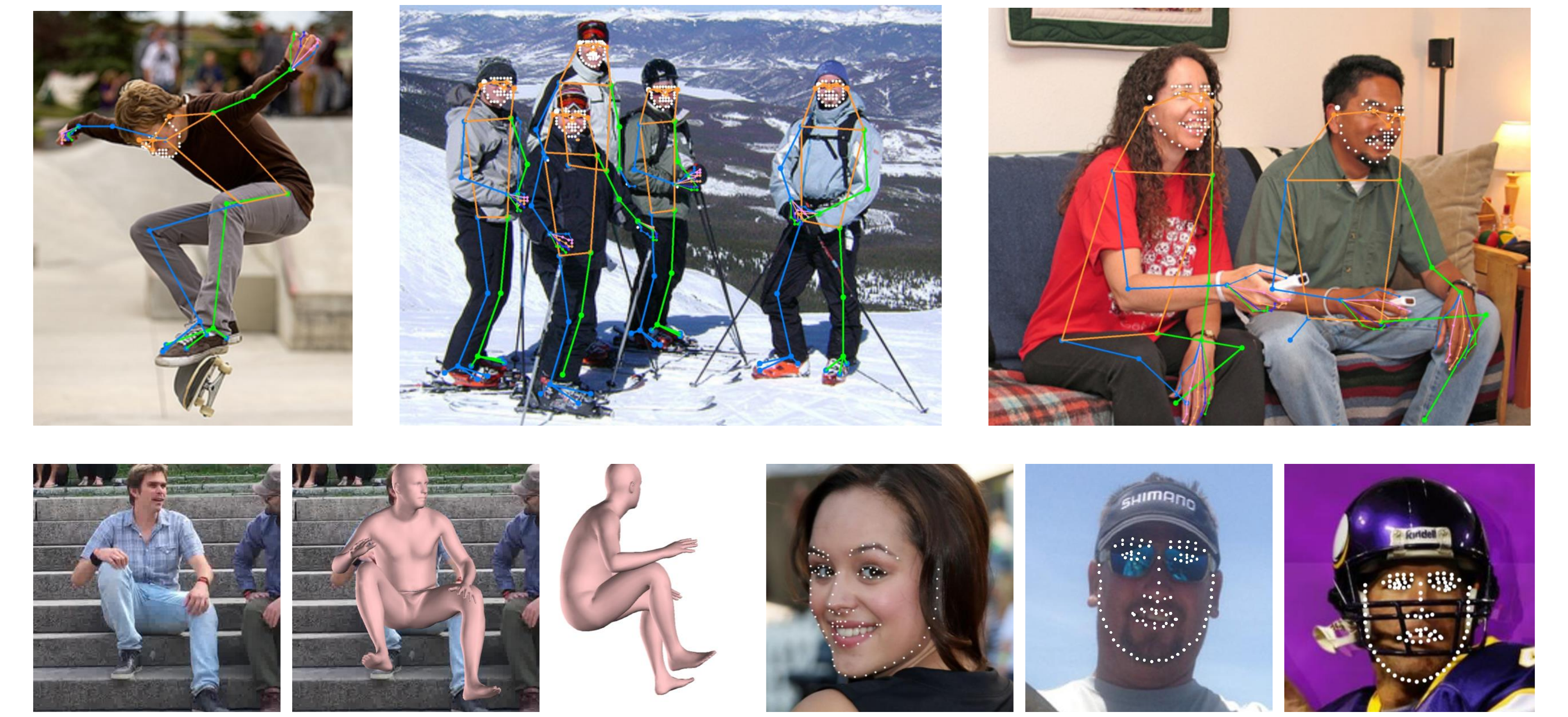}
	\vspace{-5pt}
	\caption{
	Example results of TCFormer on Whole-body pose estimation (top row), 3D mesh reconstruction (bottom left), and face alignment (bottom right). Overall, our method achieves satisfactory results on all tasks.
	}
	\label{fig:qualitative}
	\vspace{-7pt}
\end{figure*}

\textbf{Results.}
As shown in Table \ref{table:comparison_wflw_testset}, TCFormer achieves superior performance ($4.28\%$ NME), compared to the other state-of-the-art methods on the full test set and all the subsets. TCFormer even has lower error than the methods with extra information, such as PDB~\cite{feng2018wing} which uses strong data augmentation and LAB~\cite{wu2018look} which uses extra boundary information. 
The performance on face alignment validates the versatility of TCFormer beyond human body estimation.

\begin{table}[t]
    \centering
    \caption{Evaluation on ImageNet-1k \texttt{val}. 
    All results are obtained using the input size of $224\times 224$. 
    }
    \small
    \label{tab:cls}
    \setlength{\tabcolsep}{1.7mm}
    \scalebox{1.0}{
    \begin{tabular}{l|c|c|c}
        \hline
    	Method & \#Param. & FLOPs & Top-1 Acc.  \\
    	\hline
    	
        \hline
    	ResNet50~\cite{he2016deep}   & 25.6M & 4.1G & 76.1 \\
    	ResNet152~\cite{he2016deep} & 60.2M & 11.6G & 78.3 \\
        HRNet-W32~\cite{wang2020deep} & 41.2M & 8.3G & 78.5 \\
    	DeiT-Small/16~\cite{touvron2020training}  & 22.1M  & 4.6G & 79.9 \\
    	T2T-ViT$_t$-14~\cite{t2tvit} & 22.0M & 6.1G & 80.7 \\
    	HRFormer-S~\cite{yuan2021hrformer} & 13.5M & 3.6G & 81.2 \\
    	Swin-T~\cite{swin} & 29.0M & 4.5G & 81.3 \\
    	PVT-Large~~\cite{pvt} & 61.4M & 9.8G & 81.7 \\
    	TCFormer (Ours) & 25.6M & 5.9G & \textbf{82.4} \\
    	\hline
    \end{tabular}
    }
    \vspace{-5pt}
\end{table}

\subsection{Image Classification}
In order to evaluate the versatility of TCFormer on general vision tasks, we also extend it  to image classification.

\textbf{Settings.} 
We perform experiments on the ImageNet-1K dataset~\cite{russakovsky2015imagenet}. 
We apply the setting totally the same as \cite{pvt}. TCFormer without the MTA head is trained from scratch for 300 epochs with batch size 128 and evaluated on the validation set with a center crop of $224 \times 224$ patch.
For more details, please refer to the supplementary.

\textbf{Results.} 
Although the target of our model is not image classification, experimental results on ImageNet-1K show that TCFormer achieves competitive performance (82.4\% Top-1 Acc.) compared with the state-of-the-art architectures, which indicates that our CTM block also works well in extracting general image features.

\subsection{Ablation Study}
We conduct ablative analysis on the task of whole-body pose estimation as shown in Table~\ref{tab:wholebody}. 

\textbf{Effect of CTM.} 
To validate the effect of CTM (Section~\ref{sec:CTM}), we build a baseline transformer network 
by replacing CTM with a strided convolutional layer,
and find that the performance drops significantly ($-3.7\%$ AP and $-3.9\%$ AR). 
The performance drop for the parts relying on detailed information, such as the foot and the hand, is more significant than the others. The estimation AP for the foot and the hand decrease by $13.6\%$ and $5.6\%$ respectively, while the AP for the human body only drops $2.4\%$. It demonstrates that the performance drop is mainly caused by the loss of image details and validates the effectiveness of our CTM block in capturing image details in small size.

\textbf{Effect of MTA Head.} 
To validate the effect of MTA Head (Section~\ref{sec:MTA}), we replace it with a deconvolutional head~\cite{xiao2018simple} and notice a performance drop of $-1.9\%$ AP and $-1.6\%$ AR.
Especially, the performance drop on foot ($-4.0\%$ AP) and hand ($-3.6\%$ AP) is more obvious than that of the body ($-1.2\%$ AP). The results show that MTA Head is beneficial to preserving image details of body parts.

\subsection{Qualitative Evaluation}

\begin{figure}[t]
	\centering
	\includegraphics[width=0.95\linewidth]{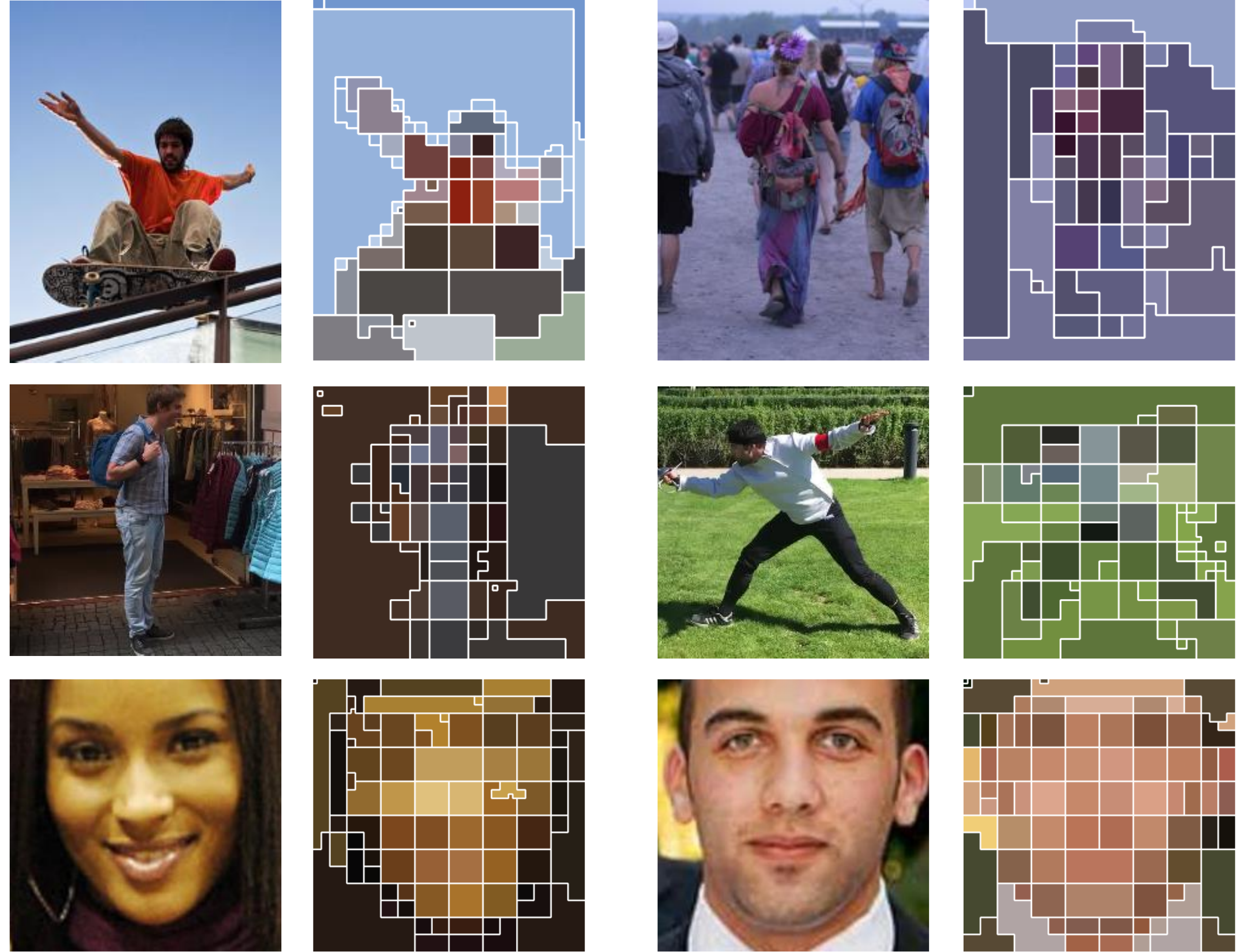}
	\vspace{-5pt}
	\caption{
	Vision tokens for whole-body pose estimation (the 1st row), 3D mesh reconstruction (the 2nd row), and face alignment (the 3rd row). For whole-body pose and 3D mesh estimation, TCFormer generates fine tokens for human face and hand. While for face alignment, fine tokens are used for face edges.
	}
	\label{fig:tokes}
	\vspace{-5pt}
\end{figure}

Figure~\ref{fig:qualitative} shows some qualitative results for human whole-body pose estimation, 3D mesh reconstruction, and face alignment. 
Figure~\ref{fig:tokes} shows some examples of the vision token distribution in the above tasks.
As shown in Figure~\ref{fig:tokes}, the vision tokens focus on the foreground regions and represent background areas with only very few tokens, even when the background is complex.
The vision tokens with fine spatial size are used for the area with important details, for example, the face and hand regions in whole-body estimation and 3D mesh reconstruction. 
For face alignment, TCFormer allocates fine tokens to the face edge regions. 
The dynamic vision token allows TCFormer to capture image details more efficiently and achieve better performance.

\section{Analysis}

\begin{figure}[t]
	\centering
	\includegraphics[width=0.95\linewidth]{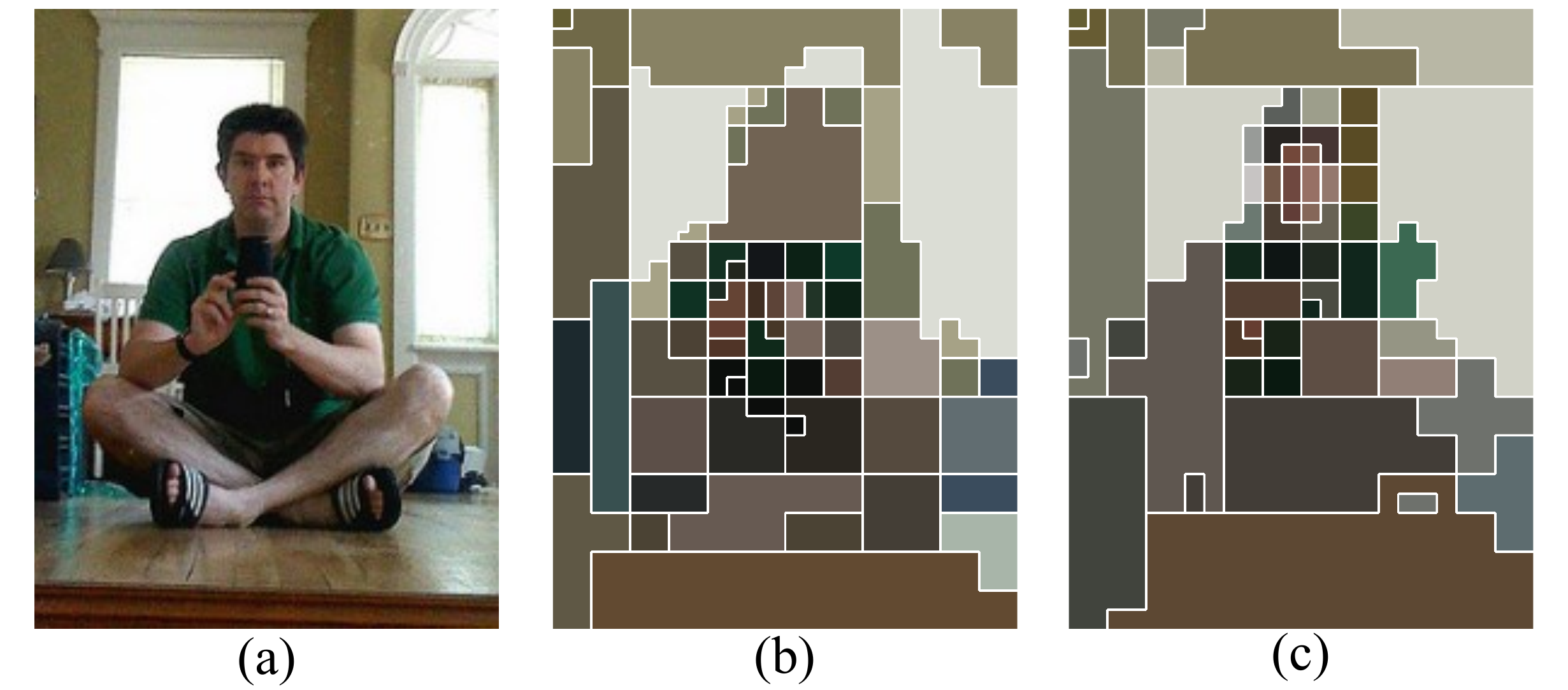}
	\vspace{-5pt}
	\caption{
	Tokens generated with different aims. We show (a) the input image and vision tokens generated by TCFormer that aims to estimate only (b) hand keypoints and (c) face keypoints. TCFormer adjusts the vision token distribution according to the task.
	}
    \vspace{-3mm}
	\label{fig:token_compare}
\end{figure}

In this part, we provide an explanation of why TCFormer focuses on regions with important details. 
Take the face region in the human whole-body pose estimation task as an example, in order to distinguish the dense keypoints on the face, TCFormer tends to learn different features for different face areas.
Since the features of the tokens representing different face parts are different, CTM block tends to group them into different clusters. So they are not merged with each other in the following merging process and are kept in a fine spatial size, which helps the feature learning in turn.

We verify the explanation by training two models with different tasks.
The first one aims to estimate only the hand keypoints, while the other one aims to estimate only the face keypoints. 
We visualize the token  distribution of these two models 
in Figure~\ref{fig:token_compare} (b) and (c), and find the token distribution to be task-specific, which matches our explanation.

\section{Conclusions and Limitations}

In this paper, we propose Token Clustering Transformer (TCFormer), a novel transformer-based architecture for human-centric vision tasks. We propose a Clustering-based Token Merge (CTM) block to equip the transformer with the capacity of preserving finer details in important regions and paying less attention to useless background information. Experiments show that our proposed method significantly improves upon its baseline model and achieves competitive performance on several human-centric vision tasks, \ie whole-body pose estimation, human mesh recovery, and face alignment.
The major limitation of TCFormer is that the computational complexity of KNN-DPC algorithm is quadratic with respect to the token number, which limits the speed of TCFormer with large input resolution. This problem can be mitigated by splitting tokens into multiple parts and performing part-wise token clustering.
We envision that the proposed method is general and can be applied to a wide range of vision tasks, \eg object detection, and semantic segmentation.
Future works will focus on exploring the effectiveness of CTM on these vision tasks.

\textbf{Acknowledgement.}
We thank Lumin Xu, Wenhai Wang, and Enze Xie for valuable discussions. 
This work is supported in part by Centre for Perceptual and Interactive Intelligence Limited, in part by the General Research Fund through the Research Grants Council of Hong Kong under Grants (Nos. 14203118, 14208619), in part by Research Impact Fund Grant No. R5001-18.
Wanli Ouyang is supported by the Australian Research Council Grant DP200103223, Australian Medical Research Future Fund MRFAI000085, and CRC-P Smart Material Recovery Facility (SMRF) – Curby Soft Plastics. 
Ping Luo is supported by the General Research Fund of HK No.27208720 and 17212120.

{\small
\bibliographystyle{ieee_fullname}
\bibliography{ref/ref.bib}
}

\clearpage

\appendix
\section*{\Large Appendix}
\setcounter{table}{0}
\renewcommand{\thetable}{A\arabic{table}}
\setcounter{figure}{0}
\renewcommand{\thefigure}{A\arabic{figure}}

\section{Detailed Settings for Image Classification}
In this section, we provide detailed experimental settings for image classification.

We train our TCFormer on the ImageNet-1K dataset~\cite{russakovsky2015imagenet}, which comprises 1.28 million training images and 50K validation images with 1,000 categories. 
We apply the data augmentations of random cropping, random flipping~\cite{szegedy2015going}, label-smoothing~\cite{szegedy2016rethinking}, Mixup~\cite{zhang2017mixup}, CutMix~\cite{yun2019cutmix}, and random erasing~\cite{zhong2020random}. 
All models are trained from scratch for 300 epochs with 8 GPUs with a batch size of 128 in each GPU.
The models are optimized with the AdamW~\cite{loshchilov2017decoupled} optimizer, with momentum of $0.9$ and weight decay of $5 \times 10^{-2}$. The initial learning rate is set to $1 \times 10^{-3}$  and decreases following the cosine schedule~\cite{loshchilov2016sgdr}. We evaluate our model on the validation set with a center crop of $224 \times 224$ patch. The experimental settings are the same as that in~\cite{pvt}.

\section{Details of TCFormer Series}
We design a series of TCFormer models with different scales for different tasks. We denote the hyper-parameters of the transformer blocks as follows and list the detailed settings of different TCFormer models in Table~\ref{tab:setting}.
 \begin{enumerate}
	\item[$\bullet$] $R_i$: The spatial reduction ratio of the transformer blocks in Stage $i$;
	\item[$\bullet$] $N_i$: The head number of the transformer blocks in Stage $i$;
	\item[$\bullet$] $E_i$: The expansion ratio of the linear layers in the transformer blocks in Stage $i$;
	\item[$\bullet$] $C_i$: The feature channel number of the vision tokens in Stage $i$.
\end{enumerate}

It's worth noting that every Clustering-based Token Merge (CTM) block contains a transformer block, whose setting is the same as the transformer blocks in the next stage.

\section{2D Whole-body Pose Estimation}
For fair comparisons with the state-of-the-art methods with larger model capacity and higher input resolution, we train TCFormer-large on the COCO-WholeBody V1.0 dataset~\cite{jin2020whole} with an input resolution of $384\times288$. Table~\ref{tab:supp_wholebody} shows the experimental results. Our TCFormer-large outperforms HRNet-w48~\cite{sun2019deep} by $1.3\%$ AP and $1.9\%$ AR, and achieves new state-of-the-art performance. Compared with other state-of-the-art methods, the gain of TCFormer is most obvious on the foot and hand, which are with small size in the input images. The results prove the capability of TCFormer in capturing details with small size.

\begin{figure}[tb]
	\centering
	\includegraphics[width=1.0\linewidth]{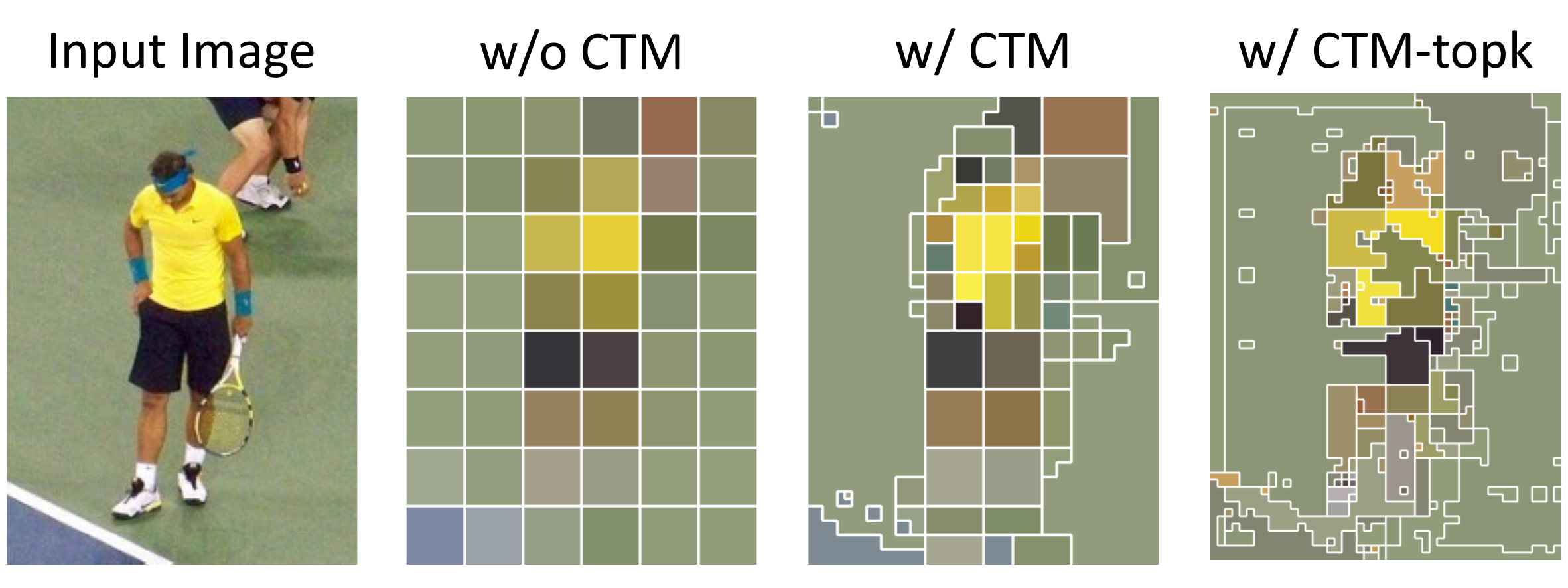}
	\caption{
	Example token distribution of models without CTM blocks, with CTM blocks, and with CTM-topk blocks. For the model with CTM-topk blocks, most vision tokens focus on a small part of the image area, leaving some human body parts represented by very few vision tokens or even merged with background tokens.
	In contrast, the vision tokens of the model with CTM blocks cover all body parts. 
	}
	\label{fig:topk}
\end{figure}

\section{More Ablation Studies}

In this section, we show the ablation study about the clustering algorithm in the CTM block. 

To validate the effect of DPC-KNN~\cite{du2016study} algorithm, we design a variant of CTM block, which determines the cluster centers by selecting the tokens with the highest importance scores and is denoted as CTM-topk block. We build a network by replacing CTM blocks in TCFormer with CTM-topk blocks and evaluate it on the task of whole-body pose estimation. 

As shown in Table~\ref{tab:ablation_topk}, replacing CTM blocks with CTM-topk blocks brings a significant performance drop of $-7.0\%$ AP and $-7.0\%$ AR. The performance of the model with CTM-topk blocks is even worse than the baseline without CTM blocks.

CTM-topk block determines the clustering centers based on the importance scores only, so most clustering centers are allocated to the regions with the highest scores. For the regions with middle scores, very few or even no clustering centers are allocated, which leads to information loss.
As shown in Figure~\ref{fig:topk}, with CTM-topk blocks, most vision tokens focus on a small part of the image area, and some body parts are represented by very few vision tokens or even merged with the background tokens, which degrades the model performance. 
In contrast, the clustering centers generated by the DPC-KNN algorithm cover all body parts, which is more suitable for human-centric vision tasks.

\section{More Qualitative Results}
\begin{figure*}[tb]
	\centering
	\includegraphics[width=0.9\linewidth]{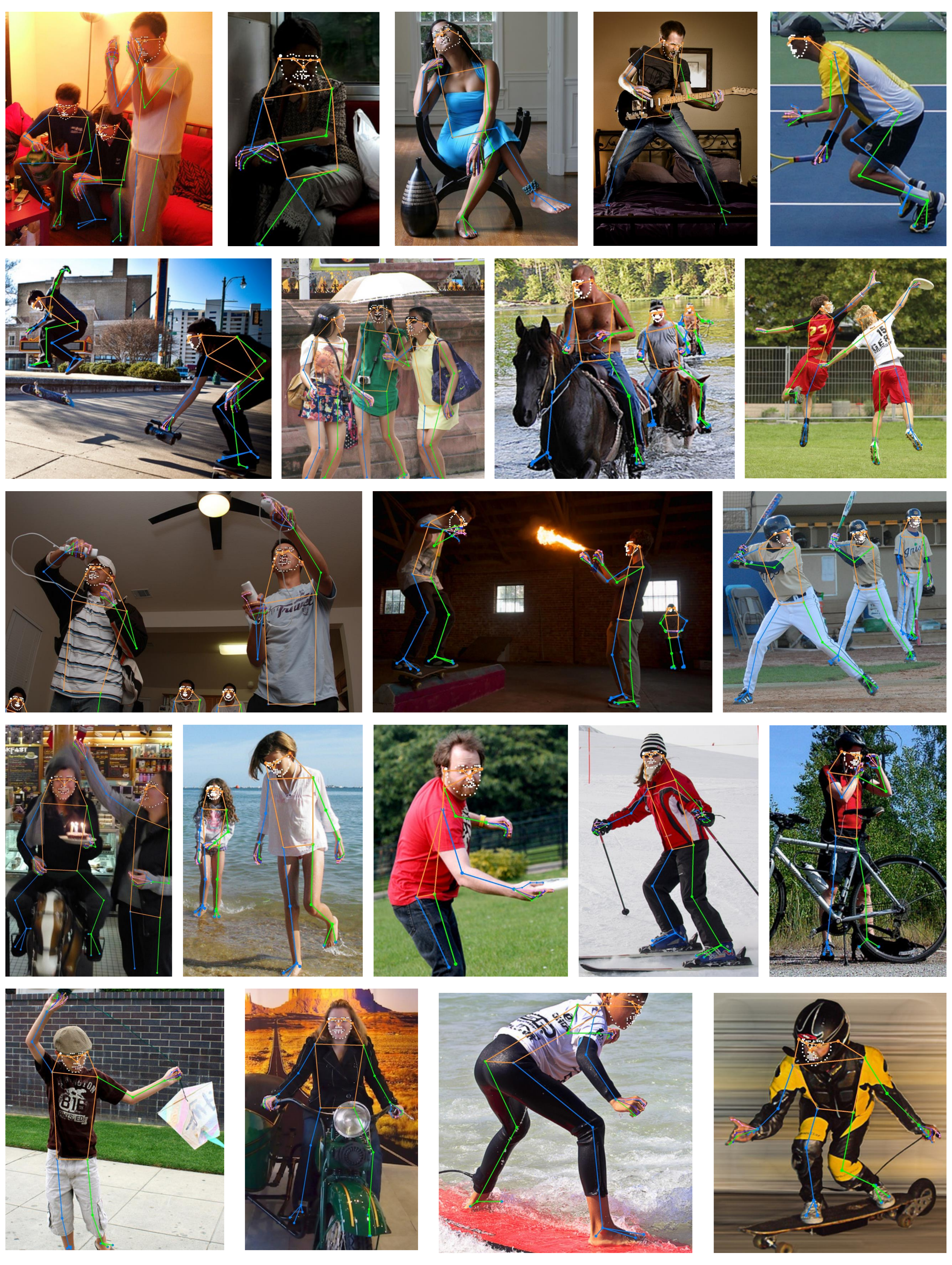}
	\caption{
	Example results of TCFormer on whole-body pose estimation.
	}
	\label{fig:wholebody_vis}
\end{figure*}

\begin{figure*}[tb]
	\centering
	\includegraphics[width=0.9\linewidth]{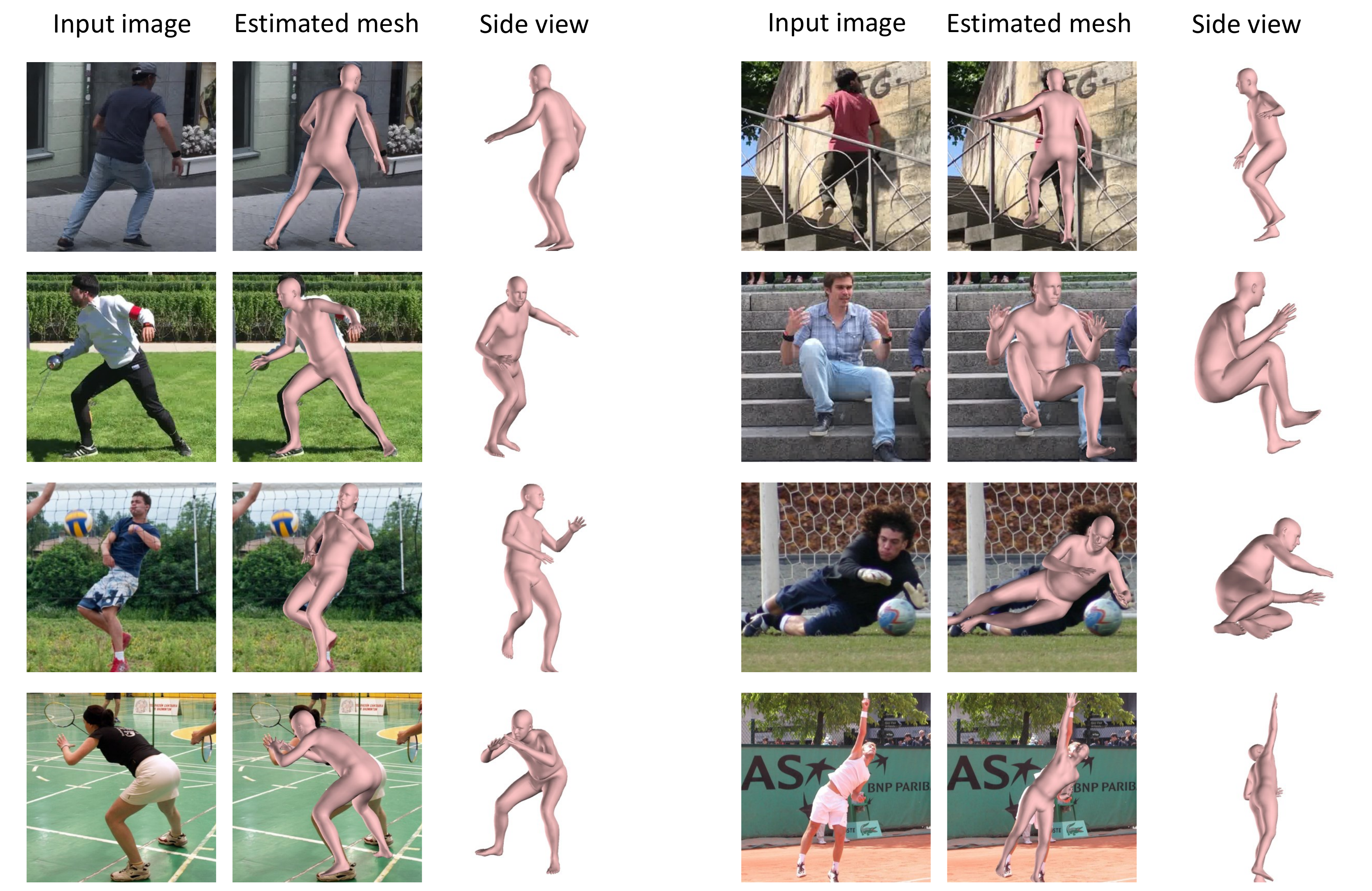}
	\caption{
	Example results of TCFormer on 3D mesh reconstruction. The top 2 rows are the results on the 3DPW~\cite{von2018recovering} dataset and the bottom 2 rows are the results on the LSP~\cite{Johnson10} test set.
	}
	\label{fig:mesh_vis}
\end{figure*}

\begin{figure*}[tb]
	\centering
	\includegraphics[width=0.9\linewidth]{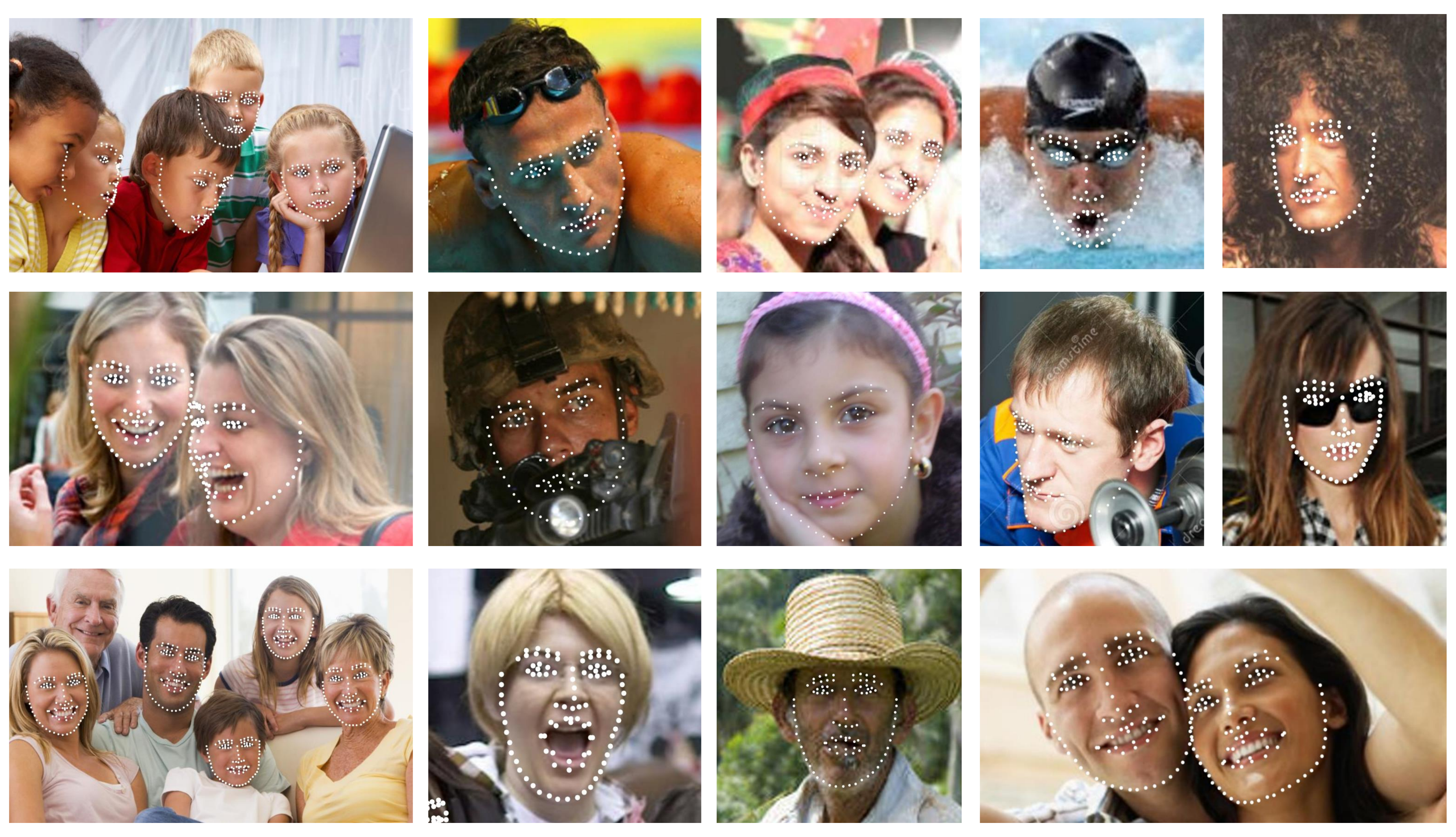}
	\caption{
	Example results of TCFormer on face alignment.
	}
	\label{fig:face_vis}
\end{figure*}

In this section, we present some qualitative results for 2D human whole-body pose estimation (Figure~\ref{fig:wholebody_vis}), 3D human mesh reconstruction (Figure~\ref{fig:mesh_vis}), and face alignment (Figure~\ref{fig:face_vis}).

As shown in Figure~\ref{fig:wholebody_vis}, our TCFormer estimates the keypoints on the hand and foot accurately, which proves the capability of TCFormer in capturing the small-scale details. Our TCFormer is also capable of handling challenges including close proximity, occlusion, and pose variation.
Figure~\ref{fig:mesh_vis} shows that our TCFormer estimates the human mesh accurately on the challenging outdoor images with large variations of background, illumination, and pose.
As shown in Figure~\ref{fig:face_vis}, TCFormer performs well on challenging cases with occlusion, heavy makeup, rare pose, and rare illumination.
Overall, the results show the robustness and versatility of our TCFormer.

\section{Visualizations about Token Distribution}
\begin{figure*}[t]
	\centering
	\includegraphics[width=0.9\linewidth]{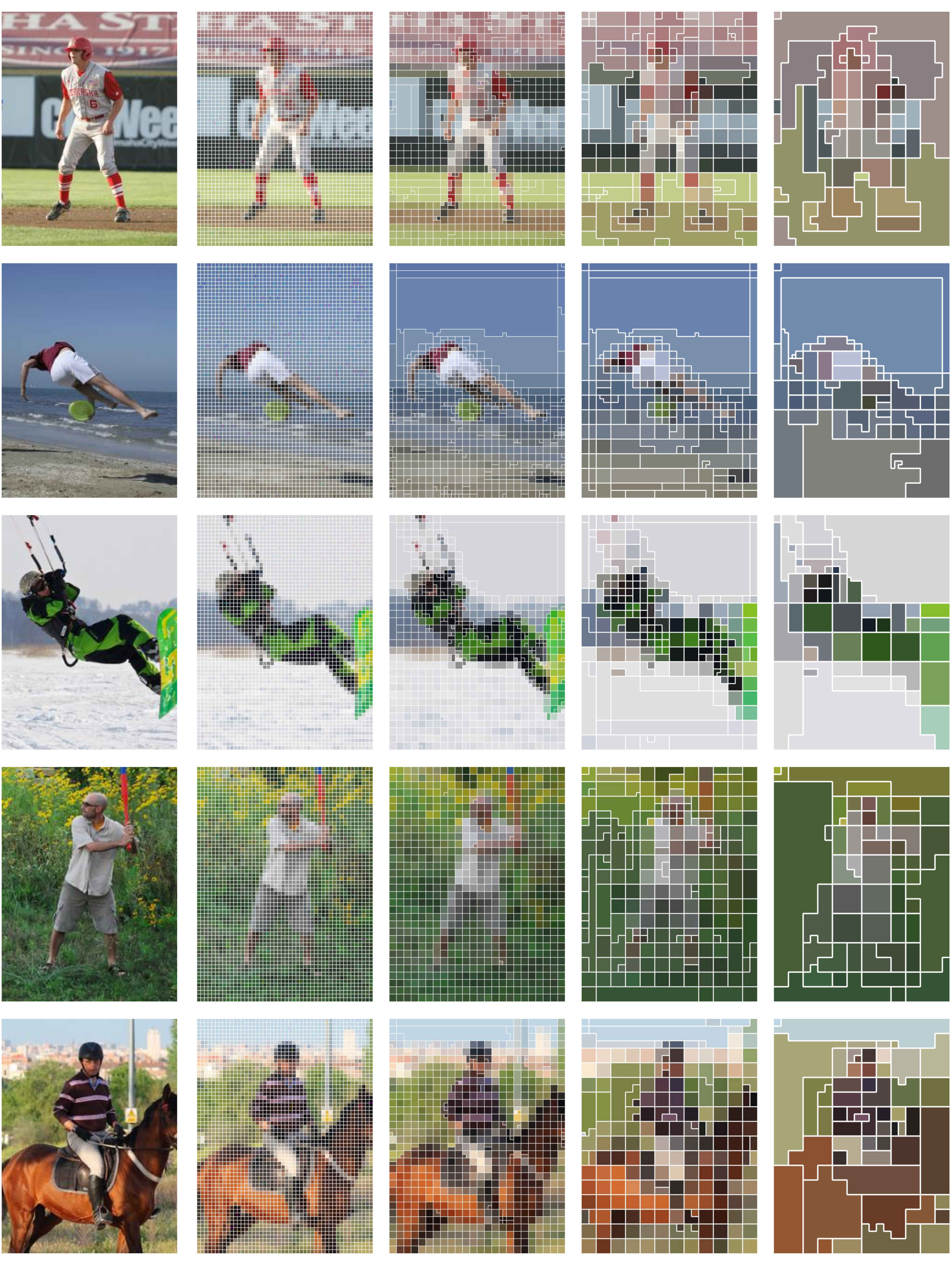}
	\caption{
	Example token distribution on 2D human whole-body pose estimation.
	}
	\label{fig:wholebody_token}
\end{figure*}

\begin{figure*}[t]
	\centering
	\includegraphics[width=0.9\linewidth]{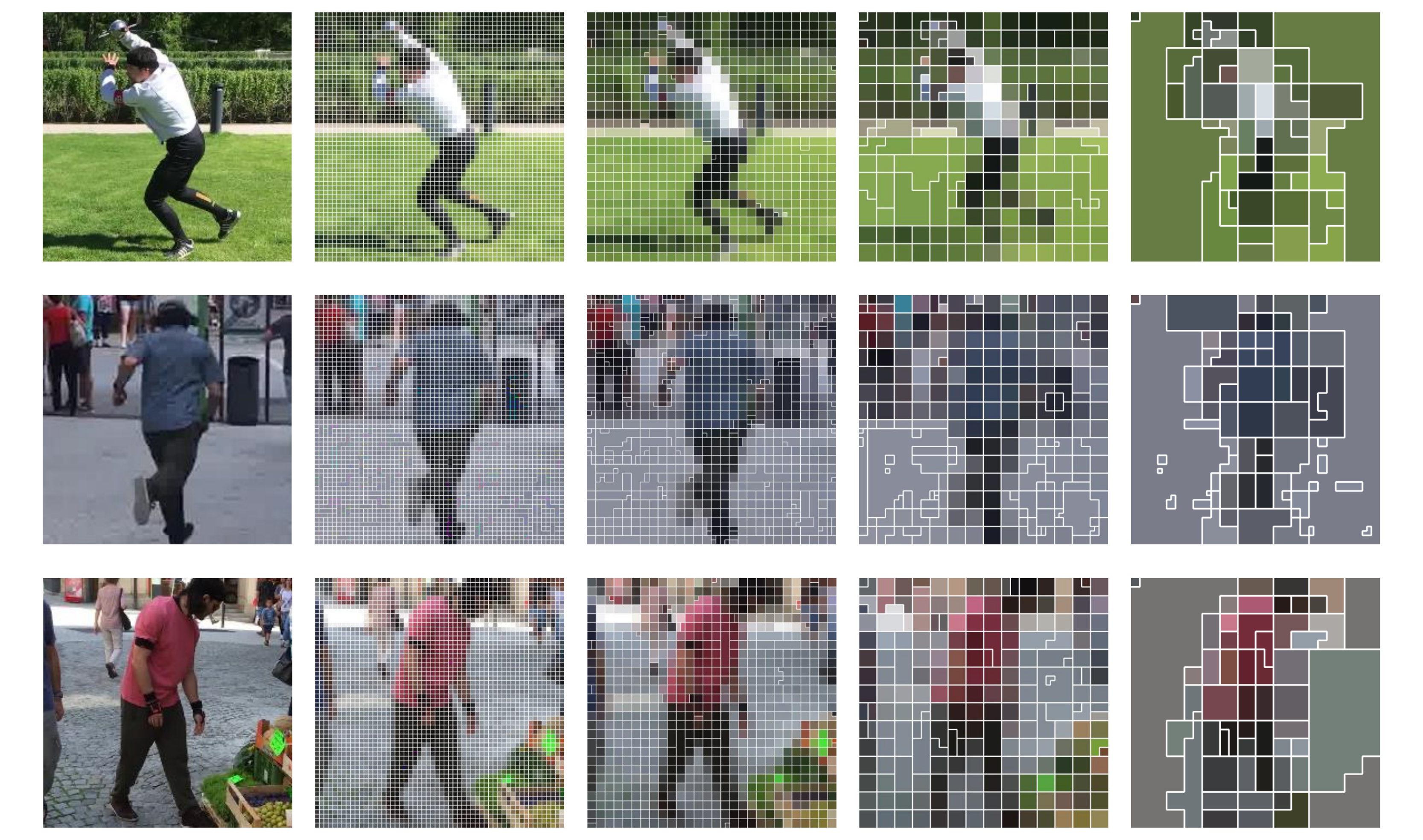}
	\caption{
	Example token distribution on 3D human mesh reconstruction.
	}
	\label{fig:mesh_token}
	\vspace{5mm}
\end{figure*}

\begin{figure*}[t]
	\centering
	\includegraphics[width=0.9\linewidth]{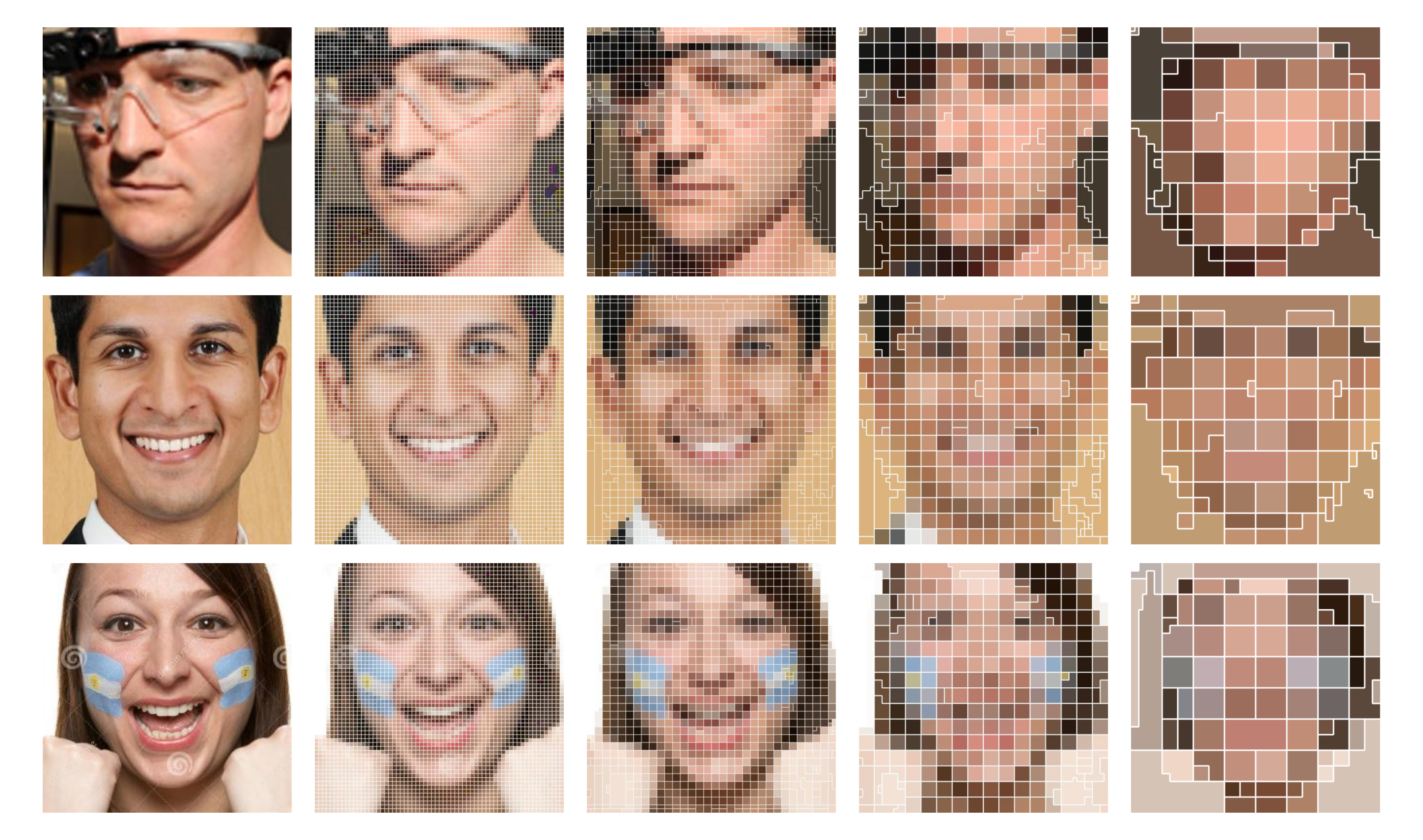}
	\caption{
	Example token distribution on face alignment.
	}
	\label{fig:face_token}
	\vspace{5mm}
\end{figure*}

\begin{figure*}[t]
	\centering
	\includegraphics[width=1.0\linewidth]{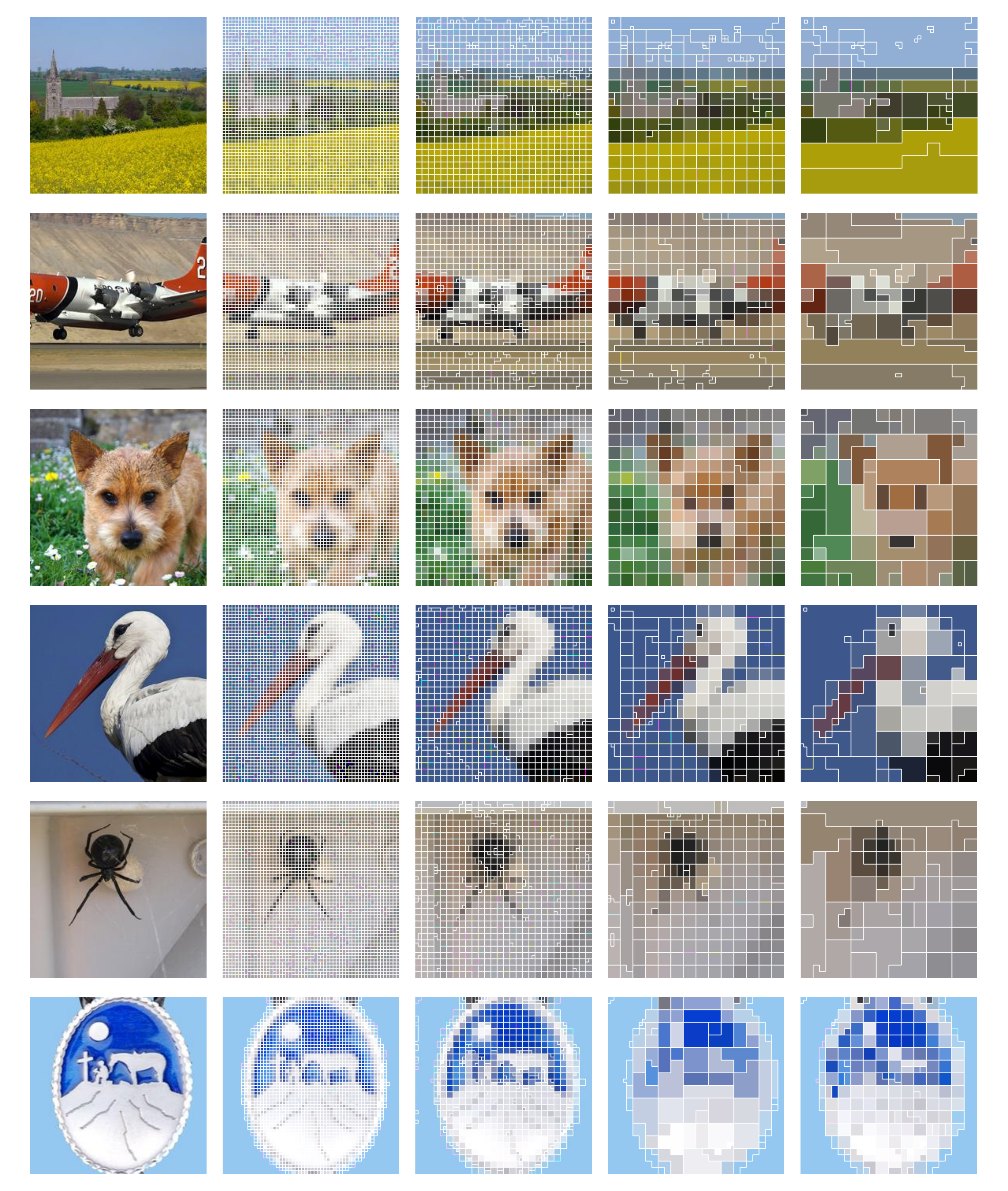}
	\caption{
	Example token distribution on image classification.
	}
	\label{fig:cls_token}
\end{figure*}

In this section, we show the vision tokens in all stages on different tasks, \ie 2D human whole-body pose estimation (Figure~\ref{fig:wholebody_token}), 3D human mesh reconstruction (Figure~\ref{fig:mesh_token}), face alignment (Figure~\ref{fig:face_token}), and image classification (Figure~\ref{fig:cls_token}).
We observe that TCFormer progressively adapts the token distribution.

As shown in Figure~\ref{fig:wholebody_token} and Figure~\ref{fig:mesh_token}, on 2D human whole-body pose estimation and human mesh estimation tasks, TCFormer merges the vision tokens of the background regions to very few tokens and pays more attention to the human body regions. For the images with simple backgrounds, such as the sky, sea, and snowfield, TCFormer merges the background tokens in stage $2$ and stage $3$. And for the images with complex backgrounds, distinguishing foreground from background requires high-level semantic features, so TCFormer merges the background vision tokens in the last stage.
On the face alignment task, TCFormer imitates the standard grid-based token distribution in the first three stages and focuses on the face edge areas in the last stage.

We can also observe targeted token distribution on image classification.
As shown in Figure~\ref{fig:cls_token}, TCFormer allocates more tokens for the informative regions and uses fewer tokens to represent the background area with little information. 
In addition, the token regions generated by TCFormer are aligned with the semantic parts.
This proves that TCFormer not only works on human-centric tasks but also on general vision tasks.

\begin{figure*}[tb]
	\centering
	\includegraphics[width=1.0\linewidth]{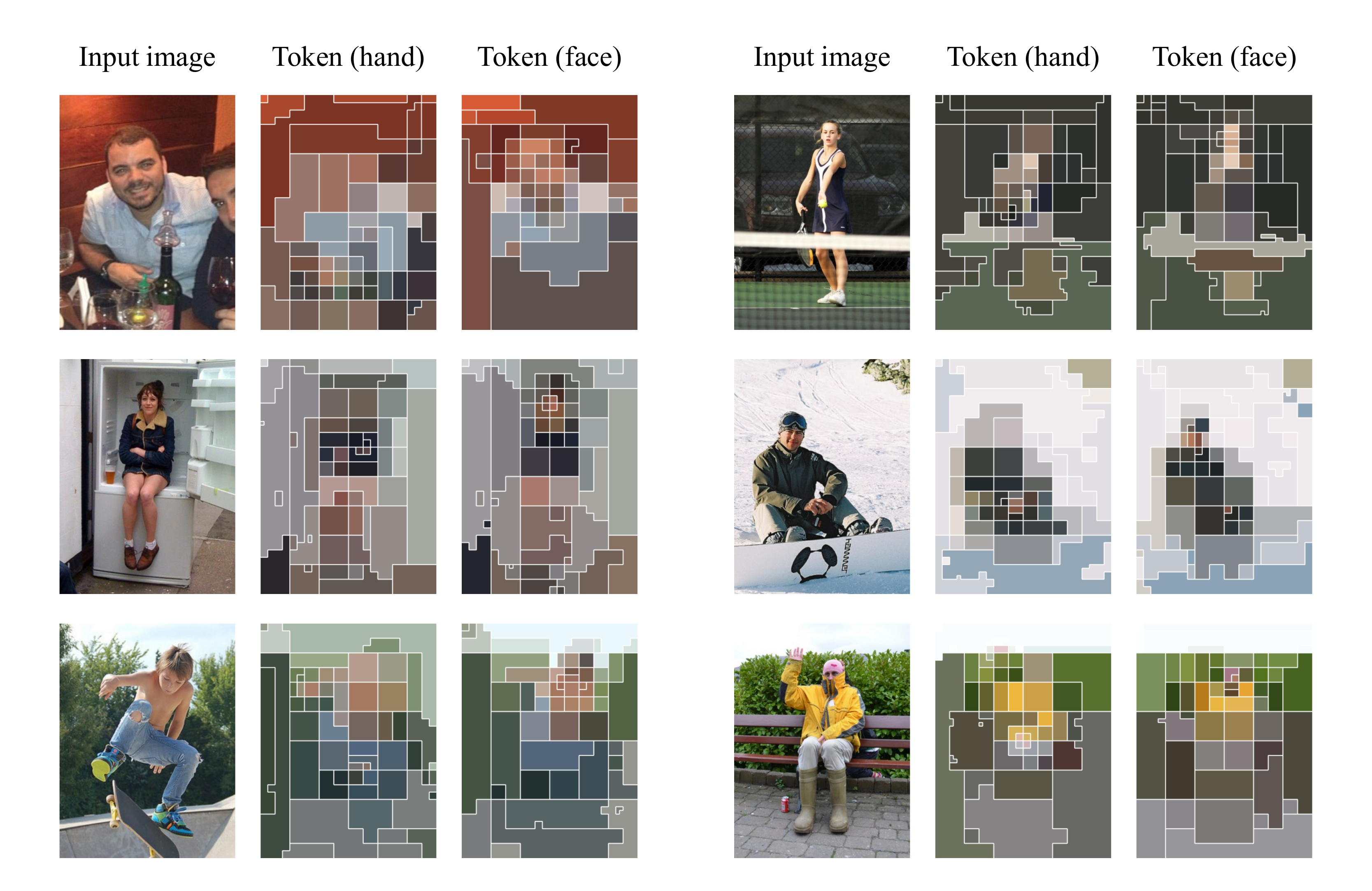}
	\caption{
	Example token distribution with different aims. 
	We show the input image, the vision tokens generated by TCFormer that aims to estimate only hand keypoints (Token (hand)) and face keypoints (Token (face)). TCFormer adjusts the vision token distribution according to the task.
	}
	\label{fig:compare_token}
\end{figure*}

We also show the distribution of tokens generated with different aims. We train two models with different tasks. The first one aims to estimate only the hand keypoints, while the other one aims to estimate only the face keypoints. 
In Figure~\ref{fig:compare_token}, we visualize the tokens generated by these two models, denoted as token (hand) and token (face) respectively. We find the token distribution to be task-specific, which proves that our TCFormer is able to focus on important image regions.

\begin{table*}[ht]
	\caption{
	Detailed settings of TCFormer series. $H$ and $W$ denotes the height and width of input images respectively.
	}
	\label{tab:setting}
	\begin{center}
    \scalebox{0.94}{
		\begin{tabular}{c|c|c|c|c|c}
		    \hline
			\multirow{2}{*}{ } & \multirow{2}{*}{Token Number} & \multirow{2}{*}{Transformer Block Setting} & \multicolumn{3}{c}{Block Number} \\
			\cline{4-6}
			& & &  TCFormer-Light & TCFormer & TCFormer-Large \\
			\hline 
			Stage1 & $\frac{H}{4} \times \frac{W}{4}$ & $\begin{array}{l}R_{1}=8,  N_{1}=1 \\ E_{1}=8,    C_1=64\end{array}$ & 2 & 3 & 3\\
			\hline 
			Stage2 & $\frac{H}{8} \times \frac{W}{8}$ & $\begin{array}{l}R_{2}=4, N_{2}=2 \\ E_{2}=8, C_{2}=128\end{array}$ & 1 & 2 & 7\\
			\hline 
			Stage3 & $\frac{H}{8} \times \frac{W}{8}$ & $\begin{array}{l}R_{3}=2, N_{3}=5 \\ E_{3}=4, C_{3}=320 \end{array}$ & 1 & 5 & 26\\
			\hline 
			Stage4 & $\frac{H}{16} \times \frac{W}{16}$ & $\begin{array}{l}R_{4}=1, N_{4}=8\\ E_{4}=4,  C_{4}=512\end{array}$ & 1 & 2 & 2\\
			\hline
		\end{tabular}
		}
	\end{center}
\end{table*}

\begin{table*}[ht]
	\caption{OKS-based Average Precision (AP) and Average Recall (AR) on the COCO-WholeBody V1.0 dataset. The baseline results are from MMPose~\cite{mmpose2020}. `*' indicates multi-scale testing.
	ZoomNet$^{\dagger}$ is trained with the COCO-WholeBody V0.5 training set.
	}
	\label{tab:supp_wholebody}
	\begin{center}
    \scalebox{0.94}{
		\begin{tabular}{c|c|cc|cc|cc|cc|aa}
			\hline
			 \multirow{2}{*}{Method} & Resolution &  \multicolumn{2}{c|}{body}  & \multicolumn{2}{c|}{foot}  & \multicolumn{2}{c|}{face}  & \multicolumn{2}{c|}{hand} & \multicolumn{2}{c}{\cellcolor{Gray}whole-body} \\
			\cline{3-12}
			& ~ &  AP     & AR     & AP   & AR     &  AP  & AR     & AP    & AR   &  AP     & AR  \\
			\hline
            SN$^{*}$~\cite{hidalgo2019single} & $480\times480$ & 0.427 & 0.583 & 0.099 & 0.369 & 0.649 & 0.697 & 0.408 & 0.580 & 0.327 & 0.456 \\ 
            OpenPose~\cite{cao2018openpose} & $480\times480$ & 0.563 & 0.612 & 0.532 & 0.645 & 0.765 & 0.840 & 0.386 & 0.433 & 0.442 & 0.523 \\ 
			PAF$^{*}$~\cite{cao2017realtime} & $480\times480$  & 0.381 & 0.526 & 0.053 & 0.278 & 0.655 & 0.701 & 0.359 & 0.528 & 0.295 & 0.405 \\ 
            AE~\cite{newell2017associative}+HRNet-w48~\cite{sun2019deep} & $512\times512$ & 0.592	& 0.686& 	0.443& 	0.595& 	0.619& 	0.674& 	0.347& 	0.438& 	0.422& 	0.532 \\ 
             HigherHRNet-w48~\cite{cheng2020higherhrnet} & $512\times512$ & 0.630 &	0.706 &	0.440 &	0.573 &	0.730 &	0.777 &	0.389 &	0.477 &	0.487 &	0.574 \\ 
            \hline
             ZoomNet$^{\dagger}$~\cite{jin2020whole} & $384\times288$ & 0.743 & 0.802 &  0.798 & 0.869 & 0.623 & 0.701 & 0.401 & 0.498 & 0.541 & 0.658 \\ 
            SBL-Res152~\cite{xiao2018simple}   & $384\times288$ & 0.703	&0.780	&0.693&	0.813&	0.751&	0.825&	0.559&	0.667&	0.610&	0.705 \\
			HRNet-w48~\cite{sun2019deep} & $384\times288$  &0.722&	0.790&	0.694&	0.799&	0.777	&0.834&	0.587	&0.679	&0.631	&0.716 \\ 
            \hline
			 TCFormer-Large (Ours) & $384\times288$ & 0.731 & 0.803 & 0.752 & 0.855 & 0.774 & 0.845 & 0.607 & 0.712 & \textbf{0.644} & \textbf{0.735} \\ 
			\hline
		\end{tabular}
		}
	\end{center}
\end{table*}

\begin{table*}[ht]
	\caption{More ablation studies on 2D human whole-body pose estimation on the COCO-WholeBody V1.0 dataset.
	}
	\label{tab:ablation_topk}
	\begin{center}
    \scalebox{0.94}{
		\begin{tabular}{c|c|cc|cc|cc|cc|aa}
			\hline
			 \multirow{2}{*}{Method} & \multirow{2}{*}{Resolution} &  \multicolumn{2}{c|}{body}  & \multicolumn{2}{c|}{foot}  & \multicolumn{2}{c|}{face}  & \multicolumn{2}{c|}{hand} & \multicolumn{2}{c}{\cellcolor{Gray}whole-body} \\
			\cline{3-12}
			& ~ &  AP     & AR     & AP   & AR     &  AP  & AR     & AP    & AR   &  AP     & AR  \\
            \hline
			 TCFormer w/o CTM  & $256\times192$ & 0.667 & 0.749&	0.562& 0.695 &	0.617 & 0.621 &	0.479& 0.590 &	0.535 & 0.639 \\ 
 			 TCFormer w/ CTM-topk  & $256\times192$ & 0.586 & 0.684 & 0.537 & 0.687 & 0.627 & 0.727 & 0.506 & 0.626 & 0.502 & 0.608 \\ 
 			 TCFormer & $256\times192$ & 0.691 & 0.770 & 0.698 & 0.813 & 0.649 & 0.746 & 0.535 & 0.650 & \textbf{0.572} & \textbf{0.678} \\ 
			\hline
		\end{tabular}
		}
	\end{center}
\end{table*}

\end{document}